\documentclass[preprint,review,12pt]{elsarticle}
\usepackage{amsmath,amsfonts}
\usepackage{amssymb}
\usepackage[ruled,vlined]{algorithm2e}
\usepackage{algpseudocode}
\usepackage{array}
\usepackage{textcomp}
\usepackage{stfloats}
\usepackage{url}
\usepackage{verbatim}
\usepackage{graphicx}
\usepackage{multirow}
\usepackage{tabu}
\usepackage{bm}
\usepackage{wrapfig}

\usepackage{color}

\newcommand{\forcond}{$k=1$ \KwTo $K$}
\journal{Pattern Recognition}

\begin{document}

\begin{frontmatter}

%% Title, authors and addresses

%% use the tnoteref command within \title for footnotes;
%% use the tnotetext command for theassociated footnote;
%% use the fnref command within \author or \address for footnotes;
%% use the fntext command for theassociated footnote;
%% use the corref command within \author for corresponding author footnotes;
%% use the cortext command for theassociated footnote;
%% use the ead command for the email address,
%% and the form \ead[url] for the home page:
%% \title{Title\tnoteref{label1}}
%% \tnotetext[label1]{}
%% \author{Name\corref{cor1}\fnref{label2}}
%% \ead{email address}
%% \ead[url]{home page}
%% \fntext[label2]{}
%% \cortext[cor1]{}
%% \affiliation{organization={},
%%             addressline={},
%%             city={},
%%             postcode={},
%%             state={},
%%             country={}}
%% \fntext[label3]{}

\title{Understanding the Vulnerability of Skeleton-based Human Activity Recognition via Black-box Attack}

%% use optional labels to link authors explicitly to addresses:
%% \author[label1,label2]{}
%% \affiliation[label1]{organization={},
%%             addressline={},
%%             city={},
%%             postcode={},
%%             state={},
%%             country={}}
%%
%% \affiliation[label2]{organization={},
%%             addressline={},
%%             city={},
%%             postcode={},
%%             state={},
%%             country={}}

\author[add1]{Yunfeng~Diao\fnref{cor2}}
\fntext[cor2]{Equal Contribution}
%\ead{diaoyunfeng@hfut.edu.cn}
\author[add2]{He~Wang\corref{cor1}\fnref{cor2}}
\cortext[cor1]{Corresponding author}
\ead{he\_wang@ucl.ac.uk}

\author[add3]{Tianjia~Shao}
\author[add4]{Yongliang~Yang}
\author[add3]{Kun~Zhou} 
\author[add5]{David~Hogg} 
\author[add1]{Meng~Wang}

\address[add1]{School of Computer Science and Information Engineering, Hefei University of Technology, Hefei, China}
\address[add2]{Department of Computer Science, University College London, London, UK}
\address[add3]{State Key Lab of CAD\&CG, Zhejiang University, Hangzhou, China}
\address[add4]{Department of Computer Science, University of Bath, Bath, UK}
\address[add5]{School of Computing, University of Leeds, Leeds, UK}

% \affiliation{organization={},%Department and Organization
%             addressline={}, 
%             city={},
%             postcode={}, 
%             state={},
%             country={}}

\begin{abstract}
%% Text of abstract
Human Activity Recognition (HAR) has been employed in a wide range of applications, e.g. self-driving cars, where safety and lives are at stake. Recently, the robustness of skeleton-based HAR methods have been questioned due to their vulnerability to adversarial attacks. However, the proposed attacks require the full-knowledge of the attacked classifier, which is overly restrictive. In this paper, we show such threats indeed exist, even when the attacker only has access to the input/output of the model. To this end, we propose the very first black-box adversarial attack approach in skeleton-based HAR called BASAR. BASAR explores the interplay between the classification boundary and the natural motion manifold. To our best knowledge, this is the first time data manifold is introduced in adversarial attacks on time series. Via BASAR, we find on-manifold adversarial samples are extremely deceitful and rather common in skeletal motions, in contrast to the common belief that adversarial samples only exist off-manifold. Through exhaustive evaluation, we show that BASAR can deliver successful attacks across classifiers, datasets, and attack modes. By attack, BASAR helps identify the potential causes of the model vulnerability and provides insights on possible improvements. Finally, to mitigate the newly identified threat, we propose a new adversarial training approach by leveraging the sophisticated distributions of on/off-manifold adversarial samples, called mixed manifold-based adversarial training (MMAT). MMAT can successfully help defend against adversarial attacks without compromising classification accuracy.
\end{abstract}

%%Graphical abstract
% \begin{graphicalabstract}
% %\includegraphics{grabs}
% \end{graphicalabstract}

%%Research highlights
% \begin{highlights}
% \item Research highlight 1
% \item Research highlight 2
% \end{highlights}

\begin{keyword}
%% keywords here, in the form: keyword \sep keyword
Black-box attack\sep skeletal action recognition\sep adversarial robustness\sep on-manifold adversarial samples
%% PACS codes here, in the form: \PACS code \sep code

%% MSC codes here, in the form: \MSC code \sep code
%% or \MSC[2008] code \sep code (2000 is the default)

\end{keyword}

\end{frontmatter}

%% \linenumbers

%% main text
\section{Introduction}
\label{sec:introduction}
Human Activity Recognition (HAR) solutions are mainly based on deep learning, which are vulnerable to adversarial attack~\cite{szegedy_intriguing_2014}. This causes major concerns especially in safety and security \cite{bountakas2023defense}, as the perturbations are \textit{imperceptible} to humans but \textit{destructive} to machine intelligence. Detecting and defending against attacks have been actively investigated \cite{bountakas2023defense}. While the research on static data (e.g. images, texts, graphs) has been widely studied, the attack on time-series data has only been recently explored ~\cite{karim_adversarial_2020}. We investigate a specific yet important type of time series data, skeletal motions, in HAR.

Skeletal motion is widely employed in HAR to mitigate issues such as lighting, occlusion, view angles, etc. Therefore, the vulnerability of skeleton-based classifiers under adversarial attack has recently drawn attention \cite{liu_adversarial_2019,zheng_towards_2020,wang_understanding_2021}. Albeit identifying a key issue that needs to be addressed, their methods are essentially \textit{white-box}. The only attempt on black-box attack is via surrogate models, i.e. attack a classifier in a white-box manner then use the results to attack the target classifier. While white-box attack requires the full knowledge of the attacked model which is overly restrictive, black-box attack via surrogate models cannot guarantee success due to its heavy dependence on the choice of the surrogate model \cite{wang_understanding_2021}. \textit{Is true black-box attack possible in skeleton-based HAR}? To answer the question, we restrict the accessible knowledge to be only the inputs/outputs of the classifiers, and propose BASAR, the \textit{very first} black-box attack method on skeleton-based activity recognition to our best knowledge.

Black-box attack on skeletal motions brings new challenges due to their unique features compared with other data. First, a skeleton usually has less than 100 Degrees of freedom (Dofs), much smaller than previously attacked data such as images/meshes. This low dimensionality leads to low-redundancy, restricting possible attacks within small subspaces. Second, \textit{imperceptibility} is a prerequisite for any successful attack, but its evaluation on skeletal motions is under-explored. Different from the attack where visual imperceptibility has high correlations with the perturbation magnitude (e.g. images), a skeletal motion has dynamics that are well-recognized by human perception. Any sparse attack, e.g. on individual joints or individual poses, albeit small, would break the dynamics and therefore be easily perceptible. In contrast, coordinated attacks on all joints and poses can provide better imperceptibility even when perturbations are relatively large \cite{wang_understanding_2021}. Consequently, the perturbation magnitude alone (as in most existing methods) is not a reliable metric for skeletal motion. Last but not least, prior methods assume that adversarial samples are off the data manifold \cite{stutz_Disentangling_2019}. As we will show, skeletal motion is one real-world example where on-manifold adversarial samples not only exist but are rather common, raising serious concerns as these on-manifold adversarial samples are \textit{implementable}.

Given a motion $\mathbf{x}$ with class label $c_\mathbf{x}$, BASAR aims to find $\mathbf{x}'$ that is close to $\mathbf{x}$ (measured by some distance function) and can fool a black-box classifier such that $c_{\mathbf{x}'}\ne c_\mathbf{x}$. BASAR formulates this process as a constrained optimization problem, aiming to find $\mathbf{x}'$ that is just outside $c_\mathbf{x}$ with a new requirement: $\mathbf{x}'$ being on the data manifold. The optimization is highly non-linear due to the compounded non-linearity of the classification boundary and the data manifold. The former dictates that any greedy search (e.g. gradient-based) near the boundary will suffer from local minima; while the latter means that not all perturbation directions result in equal visual quality (in-manifold perturbations tend to be better than off-manifold perturbations). 
% Perhaps better to make the point about the complexity of the classification boundaries in a single sentence and then add a new sentence that says 'Not all perturbation directions result in equal visual quality since in-manifold...'.  This doesn't seem to be a consequence of the complexity of the data manifold.
Consequently, there are often conflicts between these two spaces when searching for $\mathbf{x}'$. To reconcile the conflicts, we propose a new simple yet effective method called \textit{guided manifold walk} (GMW) which can compute $\mathbf{x}'$ that is close to $\mathbf{x}$ and also on the data manifold.

We exhaustively evaluate BASAR on state-of-the-art classifiers on multiple datasets in both untargeted and targeted attack tasks. The results show that not only is BASAR successful across models and datasets, it can also find on-manifold adversarial samples, in contrast to the common assumption that adversarial samples mainly exist off-manifold \cite{stutz_Disentangling_2019}. On par with recent work that also found on-manifold samples in images \cite{stutz_Disentangling_2019}, we show, for the first time, the existence and commonality of such samples in skeletal motions. We also comprehensively compare BASAR with other methods, showing the superiority of BASAR by large margins. In addition, since the perturbation magnitude alone is not enough to evaluate the attack quality, we propose a new protocol for perceptual study and conduct harsh perceptual evaluation on the naturalness, deceitfulness, and indistinguishability of the attack. The perceptual results show that on-manifold adversarial examples seem more natural and realistic than regular adversarial examples. Further motivated by this observation, we recognize that on/off-manifold adversarial examples have different distributions, which forms a new more \textit{fine-grained} description of the adversarial sample distribution. Consequently, we propose a new adversarial training approach called mixed manifold-based adversarial training (MMAT) to explore the interactions between on/off-manifold adversarial samples and clean samples during the adversarial training. We show that a proper mixture of adversarial samples with clean samples can simultaneously improve the accuracy and robustness, as opposed to the common assumption that there is always a trade-off between them~\cite{icml/Zhang/TRADES}. Overall, the philosophy behind MMAT is general and can be potentially employed on other data/tasks, e.g. images. 

%~\cite{tsipras2018robustness,icml/Zhang/TRADES,su2018robustness}
%Formally, we demonstrate that black-box adversarial attack could truly be a threat to skeleton-based human activity recognition, and systematically study their robustness towards adversaries. To this end, we 1) propose the first black-box attack method and comprehensively evaluate the vulnerability of several state-of-the-art classifiers. 2) show the existence of on-manifold adversarial samples in various datasets and demonstrate on-manifold adversarial samples are truly dangerous because they are not easily identifiable under even strict perceptual studies. 3) propose a new defense approach by leveraging the sophisticated distributions of on/off-manifold adversarial samples in data augmentation. To our best knowledge, this is the first adversarial training method in skeleton-based HAR that can simultaneously improve model robustness and accuracy. 4) propose a new perceptual study protocol to evaluate motion attack quality, addressing that there is currently no metrics suitable for evaluating motion attack quality.

%2) show the existence of on-manifold adversarial samples in various datasets and provide key insights on what classifiers tend to resist on-manifold adversarial samples. 3) show on-manifold adversarial samples are truly dangerous because they are not easily identifiable under even strict perceptual studies.

This paper is an extension of our prior research~\cite{wang_understanding_2021,diao_BASAR_2021}. The improvements and extensions include: (1) a new adversarial training method for skeleton-based HAR and detailed defense evaluation, (2) new attack experiments in more competitive classifiers and datasets, (3) new experiments integrating manifold projection with other attacks, (4) new literature review on Adversarial Defense, (5) new discussions about limitations, (6) additional details of mathematical deduction, implementation, and performance and (7) details for perceptual studies.

\section{Related Work}
\subsection{Skeleton-based Activity Recognition}

%Early research in this area has well explored a variety of hand-crafted features~\cite{Vemulapalli:2014:HAR,Fernando:2015:MVE,Devanne:2015:HAR} to recognize human actions from skeleton data. Benefiting from trained features of deep neural networks, recent  deep learning based methods have achieved the state-of-the-art performance. According to the actual skeleton data representation for learning, these methods can be categorized into sequence-based, image-based, and graph-based methods, respectively.

%Sequence-based methods treat the skeleton data as a chronological sequence of skeleton joint coordinates. The RNN architectures can in turn be employed to perform the classification~\cite{Du:2015:HRN,Liu:2016:STL,Song:2017:ESA,Zhang:2019:VAN}. Image-based methods encode the skeleton motion using a 2D pseudo-image, where one dimension represents time, and the other dimension stacks all joints of a single skeleton. As such, CNN-based image classification can be adopted~\cite{Liu:2017:ESV,Ke:2017:NRS,Kim_2017_CVPR_Workshops}. Graph-based methods consider the skeleton as a topological graph where the nodes correspond to joints and edges correspond to the bones connecting neighboring joints. Then the graph convolutional networks (GCNs) are used to effectively recognize the actions~\cite{Yan:2018:STG,Li_2019_CVPR,Shi_2019_CVPR,Shi:2019:SBA,Cheng_2020_CVPR,Zhang_2020_CVPR_2}. %

Early HAR research focuses on useful hand-crafted features. In the era of deep learning, features are automatically learned. Motions can be treated as time series of joint coordinates and modeled by Recurrent Neural Networks~\cite{si2020skeleton}. Motions can also be converted into pseudo-images and learned with Convolutional Neural Networks~\cite{li2020learning}. Graph Convolutional Networks (GCN) recently achieve state-of-the-art performance, by considering the skeleton as a graph (joints as the nodes and bones as edges)~\cite{Yan:2018:STG,Zhang_2020_CVPR_1,ctrgcn,frhead}. 
%Liu et al.~\cite{Liu_2020_CVPR} utilize disentangled multi-scale graph convolutions, which contains a unified spatial-temporal GCN operator for capturing complex spatial-temporal dependencies. Zhang et al.~\cite{Zhang_2020_CVPR_1} introduce joint semantics to the GCN model, resulting in a lightweight yet effective method.
%Zhang:2019:VAN ,Kim_2017_CVPR_Workshops Cheng_2020_CVPR,Li_2019_CVPR,
Our work is complementary to HAR, by demonstrating their vulnerability to adversarial attacks and suggesting potential improvements. We extensively evaluated BASAR on state-of-the-art methods, showing that even the very recent methods with remarkable successes are still vulnerable to adversarial attacks.

\subsection{Adversarial Attack}
%\subsubsection{White-box Attack}

Since~\cite{szegedy_intriguing_2014}, an increasing number of adversarial attack methods have been proposed in different tasks~\cite{bountakas2023defense}, such as in medical image~\cite{ma2021understanding} and person re-identification~\cite{yang2017person}. Goodfellow et al.~\cite{goodfellow_explainingah_2015} generate adversarial examples using the gradient of the model. Most of them consider the white-box setting, where the model is accessible to the attacker. Apart from common computer vision tasks such as classification~\cite{goodfellow_explainingah_2015,kurakin_advphy_2017,carlini_towards_2017}, adversarial attacks on general time series~\cite{karim_adversarial_2020} and HAR~\cite{wang_understanding_2021} have attracted attention recently. Specifically on skeleton-based HAR, an adapted version of \cite{kurakin_advphy_2017} is proposed in~\cite{liu_adversarial_2019} to attack skeletal motions. Wang et al.~\cite{wang_understanding_2021} introduced a novel perceptual loss to achieve effective and imperceptible attack. Despite existing successes, current methods are based on the full access to the attacked models, and therefore are not very applicable in real-world scenarios since the details of classifiers are not usually exposed to the attacker. 
%Hence, it is unclear whether the classifier vulnerability under (white-box) adversarial attack is a real threat. We on the other hand show that black-box attack is possible and is more threatening.
%wei2019sparse,
%~\cite{karim_adversarial_2020,fawaz_adversarial_2019,harford2020adversarial}
%Tanaka et al.\cite{tanaka2022adversarial} suggest only perturbing the lengths of bones to fool skeleton-based HAR.

%\subsubsection{Black-box Attack}
The difficulties of white-box attack in the real-world motivate the black-box attack, where attackers cannot access the full information of the attacked model. A simple approach is transfer-based attack, which generates adversarial samples from one surrogate model via white-box attack~\cite{wang_understanding_2021}. Existing black-box methods on skeletal motions~\cite{liu_adversarial_2019, wang_understanding_2021} all rely on such a method, and cannot guarantee success due to the heavy dependence on the surrogate model~\cite{wang_understanding_2021}. In a truly black-box setting, only the final class labels (hard-labels) can be used, such setting is also called hard-label attack. Brendel et al. \cite{drendel_decisionaa_2018} perform the first hard-label attack by a random walk along the decision boundary. The Rays attack~\cite{Rays} employs a discrete search algorithm to reduce unnecessary searches. However, existing hard-label attacks do not explicitly model the data manifold, and are hence incapable of considering the visual imperceptibility of the attack if they are adapted to attack skeletal motions.

\subsection{Adversarial Training}
The original idea of adversarial training (AT)~\cite{szegedy_intriguing_2014} is to train classifiers with a mixture of adversarial samples and clean data, to defend against adversarial attacks. Goodfellow et al.~\cite{goodfellow_explainingah_2015} further extended the approach by using an attacker to generate adversarial examples during AT. Madry et al. ~\cite{DBLP:conf/iclr/MadryMSTV18} later redefined AT using robust optimization. Despite the significant progresses in AT~\cite{bountakas2023defense,wang2020improving}, existing methods all compromise the accuracy to different extents. More importantly, the defense for skeleton-based HAR has still been largely under-explored. Early research~\cite{icml/Zhang/TRADES} postulates that the trade-off between adversarial robustness and accuracy may be inherent. However, some recent works have proven that the trade-off can be mitigated or even theoretically eliminated. A series of works~\cite{DBLP:conf/icml/WangPDL0Y23,addepalli2022efficient} have demonstrated that using extra (synthesis) data can mitigate such a trade-off. Stutz et al.~\cite{cvpr/Stutz/onm} showed the existence of on-manifold adversarial samples, and reckon that on-manifold robustness is essentially related to the model generalization. Yang et al. \cite{nips/Yang} found if different classes are at least $2r$ apart, then there exists an ideal classifier which can defend against any attacks bounded by $r$ without compromising the accuracy. Pang et al. ~\cite{pang2022robustness} attribute the trade-off to the improper definition of robustness, hence they substitute inductive bias of local invariance with local equivariance to redefine the robust error. Very recently, adversarial defenses for video modality have just been attempted~\cite{lo2021defending,yang2022video}. In this paper, we further extend~\cite{wang_understanding_2021,diao_BASAR_2021} to propose a new on-manifold adversarial training for skeleton-based HAR. The results show the proposed defense can potentially truly eliminate the trade-off between robustness and accuracy for skeletal motions.
%Early research~\cite{icml/Zhang/TRADES} postulates that the trade-off between adversarial robustness and accuracy may be inherent. However, some recent works have proven that the trade-off can be mitigated or even theoretically eliminated. Stutz et al.~\cite{cvpr/Stutz/onm} showed the existence of on-manifold adversarial samples, and reckon that on-manifold robustness is essentially related to the model generalization. Yang et al. \cite{nips/Yang} found if different classes are at least $2r$ apart, then there exists an ideal classifier which can defend against any attacks bounded by $r$ without compromising the accuracy.
%\cite{icml/Zhang/TRADES,pami/Miyato_VAT,icml/FAT,ijcai/CAT,zhang2021geometryaware,nips/Wu/AWP,nips/CarmonRSDL19,augustin2020adversarial,rice2020overfitting,Wang2020MART}
%Semi-Supervised Learning-based methods can utilize extra data~\cite{pami/Miyato_VAT} to mitigate the problem, and infinite data will improve the trade-off~\cite{icml/RaghunathanXYDL20}.

% Considering that the defense for skeleton-based HAR has been largely under-explored, we further extend~\cite{wang_understanding_2021,diao_BASAR_2021} to propose a new on-manifold adversarial training for skeleton-based HAR, with thorough evaluation and comparisons with other defense methods. The results show the proposed defense can potentially truly eliminate the trade-off between robustness and accuracy for skeletal motions.

\section{Methodology}
We denote a motion with $n$ poses as $\mathbf{x}$ = $\{\textbf{x}^1,\dots, \textbf{x}^n\}$, where each pose $\textbf{x}^t$ = $\{q_1,\dots,q_m\}$ includes $m$ Dofs (joint positions or angles). A trained activity classifier $G$ maps a motion to a probabilistic distribution over classes, $G$: $\mathbf{x}\rightarrow \mathbb{R}^C$ where $C$ is the total number of action classes. The class label $c_\mathbf{x}$ then can be derived e.g. via \textit{softmax}. An adversarial sample $\mathbf{x'}$ corresponding to $\mathbf{x}$ can be found via~\cite{carlini_towards_2017}:
\begin{align}
     \text{minimize }\ \ L(\mathbf{x}, \mathbf{x'})\ \  \text{ s.t. }\ \ c_{\mathbf{x'}} = c \text{ and } \mathbf{x'} \in [0, 1]^{m\times n}
\end{align}
where $L$ is the Euclidean distance. $c$ is the targeted class. Note that the constraint can also be replaced by $c_{\mathbf{x'}} \ne c_{\mathbf{x}}$ for untargeted attack. However, simply applying this attack to skeletal motions is not sufficient because it only restricts the adversarial sample $\mathbf{x'}$ in a hyper-cube $[0,1]^{m\times n}$. Given that human poses lie in a natural pose manifold $\mathcal{M}$, $\mathbf{x'}$ can easily contain off-manifold poses which are unnatural/implausible and easily perceptible. We therefore add another constraint $\mathbf{x'} \in \mathcal{M}$:
\begin{align}
\label{eq:opt}
     & \text{minimize }\ \ L(\mathbf{x}, \mathbf{x'})\ \  \text{ s.t. }\ \   \mathbf{x'} \in [0, 1]^{m\times n}, \mathbf{x'} \in \mathcal{M} \nonumber \\
    & c_{\mathbf{x'}} = c\text{ (targeted)} \text{ or } c_{\mathbf{x'}} \ne c_{\mathbf{x}}\text{ (untargeted)}. 
\end{align}
In practice, we find that $\mathbf{x'} \in [0, 1]^{m\times n}$ is less restrictive than other constraints and always satisfied. The optimization is highly nonlinear and cannot be solved analytically. It thus requires a numerical solution. 

\begin{figure}[t]
    \centering
    \includegraphics[width=0.7\linewidth]{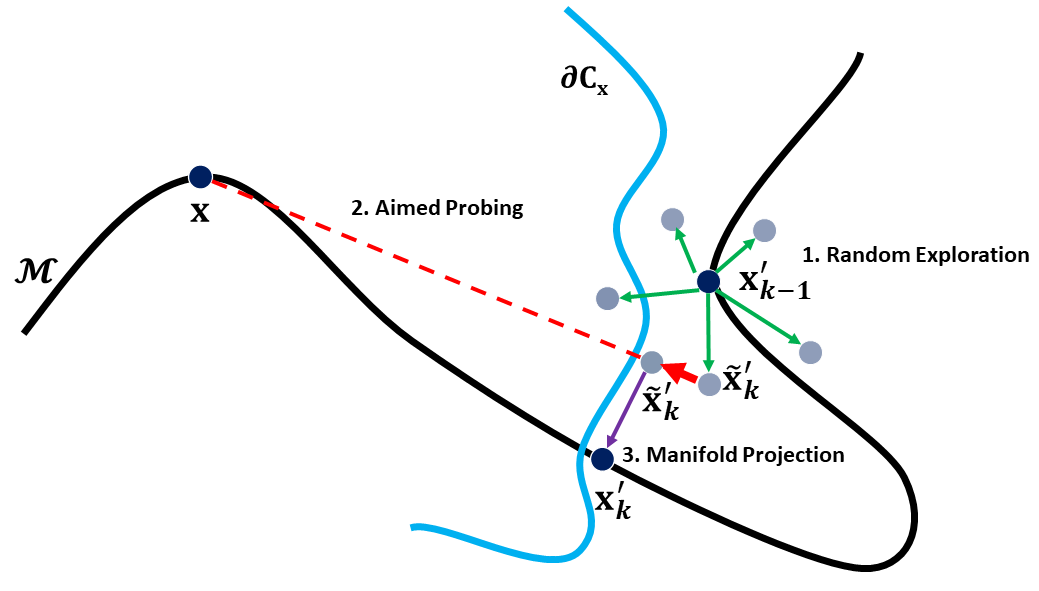}
    \caption{An abstract 2D illustration of BASAR. $\textbf{x}$ is the attacked motion. $\textbf{x}'_k$ is the ideal adversarial sample in iteration $k$. $\mathcal{M}$ (black line) is the natural pose manifold and $\partial C_{\mathbf{x}}$ (blue line) is the class boundary of $c_{\mathbf{x}}$. $\textbf{x}'_{k-1}$ is the result of last iteration. $\Tilde{\textbf{x}}'_k$ is the intermediate result of the current iteration.}
    \label{fig:overall}
\end{figure}

\subsection{Guided Manifold Walk}
We propose a new method called \textit{Guided Manifold Walk} (GMW) to solve Equation~\ref{eq:opt}. For simplicity, we start with an abstract 2D illustration of GMW on $\mathbf{x}$ shown in Figure~\ref{fig:overall}. $\mathbf{x}'_k$ is the ideal adversarial sample which is on-manifold and close to $\mathbf{x}$ in the $k$th iteration.  Given the non-linearity of the classification boundary and the data manifold, BASAR aims to exploit the properties of both simultaneously. The GMW is an iterative approach where two major steps are alternatively conducted. One step is to find an adversarial sample that is close to $\mathbf{x}$ and the other one is to find the closest sample on the data manifold from an arbitrary off-manifold position. Since the former mainly considers the classification boundary, we design two sub-routines: \textit{random exploration} and \textit{aimed probing}. Similar sampling strategies have been attempted in attacking images~\cite{drendel_decisionaa_2018}. We extend them to motions by treating a whole motion as an $n \times m$ image, with joint weighting and automatically thresholded attacks. Random exploration is to explore the vicinity of the current adversarial sample to find a random sample (step 1 in Figure~\ref{fig:overall}). Aimed probing is to find a sample in proximity to $\partial C_{\mathbf{x}}$ and is closer to $\mathbf{x}$ (step 2 in Figure~\ref{fig:overall}). The details are given in Section~\ref{sec:re} and \ref{sec:ap}. Finally, we design a sub-routine: \textit{manifold projection} which is to project an off-manifold sample $\Tilde{\mathbf{x}}'_k$ onto $\mathcal{M}$ to obtain $\mathbf{x}'_k$  (step 3 in Figure~\ref{fig:overall}). This is one key element of our approach in bringing the motion manifold into adversarial attack. The algorithm overview is given in Algorithm~\ref{alg:al1}, where $\lambda$ and $\beta$ are hyper-parameters and $l$ is a distance function. Next, we give details of all sub-routines.
%algorithm 1
\begin{algorithm}[!htb]
\scriptsize
%\footnotesize
\SetAlgoLined
\label{alg:al1}
\textbf{Input}: $\mathbf{x}$: attacked motion\; 
$\widetilde{\mathbf{x}}_0$: random sample, where \\ \ \ \ \ \ \ $c_{\widetilde{\mathbf{x}}_0} = c\text{ (targeted)}
 \text{ or } c_{\widetilde{\mathbf{x}}_0} \ne c_{\mathbf{x}}\text{ (untargeted)}$\; 
 $K$: maximum number of iterations; $\epsilon$: perturbation threshold; $\lambda$, $\beta_1$ and $\beta_2$: hyper-parameters;
 
\textbf{Initialization}: $\mathbf{x}'_0 = Aimed Probing (\widetilde{\mathbf{x}}_0 , \mathbf{x},\beta_1)$, so that $\mathbf{x}'_0$ is adversarial\ and $\beta_1$ = update($\beta_1$, $\mathbf{x}'_0$)\; %and $\Tilde{\mathbf{x}}'_1 = \mathbf{x}'_0$\;
 \For{\forcond}{
    $\widetilde{\mathbf{x}}_k = Random Exploration (\mathbf{x}'_{k-1}, \mathbf{x}, \lambda)$\;
    \While{$\widetilde{\mathbf{x}}_k$ is not adversarial \text{ and } $\lambda \ge 10^{-10}$ }{
        $\lambda$ = update($\lambda$, $\widetilde{\mathbf{x}}_k$)\;
        $\widetilde{\mathbf{x}}_k = Random Exploration (\mathbf{x}'_{k-1}, \mathbf{x}, \lambda)$\;
    }
    \leIf{$\lambda \ge 10^{-10}$}{$\mathbf{x}'_k = \widetilde{\mathbf{x}}_k$; $\lambda$ = update($\lambda$, $\widetilde{\mathbf{x}}_k$)\;}{$\mathbf{x}'_k=\mathbf{x}'_{k-1}$; break}
    $\widetilde{\mathbf{x}}_k = Aimed Probing (\mathbf{x}'_k, \mathbf{x}, \beta_1)$\;
    \While{$\widetilde{\mathbf{x}}_k$ is not adversarial \text{ and } $\beta_1 \ge 10^{-10}$ }{
        $\beta_1$ = update($\beta_1$, $\widetilde{\mathbf{x}}_k$)\;
        $\widetilde{\mathbf{x}}_k = Aimed Probing (\mathbf{x}'_k, \mathbf{x}, \beta_1)$\;
        %\lIf{$\widetilde{\mathbf{x}}_k$ is adversarial}{break}
                }
    \leIf{$\beta_1 \ge 10^{-10}$}{$\mathbf{x}'_k = \widetilde{\mathbf{x}}_k$; $\beta_1$ = update($\beta_1$, $\widetilde{\mathbf{x}}_k$)\;}{break}
    $\widetilde{\mathbf{x}}_k = Manifold Projection (\mathbf{x}'_k, \mathbf{x})$; $\hat{\mathbf{x}}_k = \widetilde{\mathbf{x}}_k $\;
    \While{$\widetilde{\mathbf{x}}_k$ is not adversarial \text{ and } $\beta_2 \ge 10^{-10}$}{
      $\beta_2$ = update($\beta_2$, $\hat{\mathbf{x}}_k$)\;
      $\widetilde{\mathbf{x}}_k = Aimed Probing (\mathbf{x}'_k, \hat{\mathbf{x}}_k, \beta_2)$\;
    }
    \leIf{$\beta_2 \ge 10^{-10}$}{$\mathbf{x}'_k = \widetilde{\mathbf{x}}_k$;
    $\beta_2$ = update($\beta_2$, $\widetilde{\mathbf{x}}_k$)\;}{break}
    \lIf{l($\mathbf{x}'_k$,$\mathbf{x}$) $< \epsilon$}{break}
}
\Return $\mathbf{x}'_k$\;
 \caption{Overview of the GMW}
\end{algorithm}

\subsection{Random Exploration} 
\label{sec:re}
Random exploration is to explore in proximity to the classification boundary, by making a small step towards a random direction: 
\begin{align}
\label{eq:re}
& \widetilde{\mathbf{x}} =\mathbf{x}' + \mathbf{W}\Delta, \text{where }\Delta = \mathbf{R} - (\mathbf{R}^T\mathbf{d})\mathbf{d}, \  \mathbf{d} = \frac{\mathbf{x} - \mathbf{x}'}{\| \mathbf{x}- \mathbf{x}'\|}, \nonumber \\
& \mathbf{R} = \lambda \frac{\mathbf{r}}{\| \mathbf{r}\|} \| \mathbf{x} - \mathbf{x}'\|\text{, } r \in N(0, \mathbf{I}),
\end{align}
% & \widetilde{\mathbf{x}} =\mathbf{x}' + \mathbf{W}\Delta, \nonumber \\
% & \text{where }\Delta _{*} = \mathbf{R}_{*} - (\mathbf{R}_{*}^T\mathbf{d}_{*})\mathbf{d}_{*}, \  \mathbf{d}_{*} = \frac{\mathbf{x}_{*} - \mathbf{x}'_{*}}{\| \mathbf{x}_{*} - \mathbf{x}'_{*}\|}, \nonumber \\
% & \mathbf{R}_{*} = \lambda \frac{\mathbf{r}}{\| \mathbf{r}\|} \| \mathbf{x}_{*} - \mathbf{x}'_{*}\|\text{, } r \in N(0, \mathbf{I}),
% \end{align}
% \begin{equation}
% \label{eq:re}
% \widetilde{\mathbf{x}} =\mathbf{x}' + \mathbf{W}\Delta, 
% \end{equation}
% $ \text{where }\Delta _{*} = \mathbf{R}_{*} - (\mathbf{R}_{*}^T\mathbf{d}_{*})\mathbf{d}_{*}, \  \mathbf{d}_{*} = \frac{\mathbf{x}_{*} - \mathbf{x}'_{*}}{\| \mathbf{x}_{*} - \mathbf{x}'_{*}\|}, \nonumber \\
%  \mathbf{R}_{*} = \lambda \frac{\mathbf{r}}{\| \mathbf{r}\|} \| \mathbf{x}_{*} - \mathbf{x}'_{*}\|\text{, } r \in N(0, \mathbf{I}),$
where $\widetilde{\mathbf{x}}$ is the new perturbed sample, $\mathbf{x}$ and $\mathbf{x}'$ are the attacked motion and current adversarial sample. The perturbation on $\mathbf{x}'$, $\Delta$, is weighted by $\mathbf{W}$ - a diagonal matrix with joint weights. This is based on the experiments and perceptual study which suggest that equal perturbations on different joints are not equally effective and imperceptible, e.g. perturbations on the spinal joints cause larger visual distortion but are less effective in attacks. We therefore weight them differently. $\Delta$ controls the direction and magnitude of the perturbation, and depends on two variables $\mathbf{R}$ and $\mathbf{d}$. $\mathbf{d}$ is the directional vector from $\mathbf{x}'$ to $\mathbf{x}$. $\mathbf{R}$ is a random directional vector sampled from a Normal distribution $N(0, \mathbf{I})$ where $\mathbf{I}$ is an identity matrix, $\mathbf{I} \in R^{z \times z}$, $z = mn$, $m$ is the number of Dofs in one frame and $n$ is total frame number. This directional vector is scaled by $\| \mathbf{x} - \mathbf{x}'\|$ and $\lambda$.

%In Equation~\ref{eq:re}, $\Delta$ is \textit{orthogonal} to $\mathbf{d}$. The random exploration essentially explores in the directions that are orthogonal to the direction towards $\mathbf{x}$, within the local area around $\mathbf{x}'$: $\mathbf{x}' + \mathbf{W} \Delta$. This mainly aims to avoid local minima along the $\partial C_{\mathbf{x}}$ when approaching $\mathbf{x}$. Also, we introduce adaptive attack (update($\lambda$, $\widetilde{\mathbf{x}}$ in Algorithm~\ref{alg:al1}). If $\widetilde{\mathbf{x}}$ is not adversarial, we decrease $\lambda$ and perform the random exploration again; otherwise, we increase $\lambda$, then enter the next sub-routine. A similar perturbation strategy to Equation~\ref{eq:re} is employed in~\cite{drendel_decisionaa_2018} for images, but we adapt it for skeletal motions with joint weights and adaptive attack. More details can be found in supplementary material.

\subsection{Aimed Probing}
\label{sec:ap}
Aimed probing is straightforward, aiming to find a new adversarial sample between the perturbed motion and the original, so that the new sample is closer to the attacked motion and remains adversarial: $\widetilde{\mathbf{x}} = \mathbf{x}' + \beta(\mathbf{x} - \mathbf{x}')$, where $\beta$ is a forward step size. Similar to $\lambda$, $\beta$ is decreased to conduct the aimed probing again if $\widetilde{\mathbf{x}}$ is not adversarial; otherwise, we increase $\beta$, then enter the next sub-routine. %{\color{red} Different from \cite{drendel_decisionaa_2018}, we dynamically control $\beta$ which is decreased to conduct the aimed probing again if $\widetilde{\mathbf{x}}$ is not adversarial; otherwise, we increase $\beta$, then enter the next sub-routine.}

\subsection{Manifold Projection}
After aimed probing and random exploration, the perturbed motion $\widetilde{\mathbf{x}}$ is often off the manifold, resulting in implausible and unnatural poses. We thus project them back to the manifold. The natural pose manifold can be regarded as poses that do not violate bone lengths or joint limits, i.e. they are realizable by humans. Further, a motion is regarded as on-manifold if all its poses are on-manifold. The motion manifold can be obtained in two ways: explicit modeling  or implicit learning. Using implicit learning would require to train a data-driven model then use it for projection, breaking BASAR into a two-step system. Therefore we employ explicit modeling. Specifically, we replace the constraint $\mathbf{x}' \in \mathcal{M}$ in Equation~\ref{eq:opt} with hard constraints on bone lengths and joint limits. We also constrain the dynamics of $\mathbf{x}'$ to be similar to the original motion $\mathbf{x}$:
\begin{align}
\label{eq:gmw_Euclidean}
    & \min_\mathbf{x'}\ \ L(\widetilde{\mathbf{x}}, \mathbf{x'})  + w L(\mathbf{\ddot x}, \mathbf{\ddot x'})  \nonumber \\ 
    & \text{s.t.}\ \ B'_{i} = B_i \text{ and } \theta_j^{\min} \le \theta'_{j} \le \theta_j^{\max} \nonumber \\
    & C_{\mathbf{x'}} = c\text{ (targeted)} \text{ or } C_{\mathbf{x'}} \ne C_{\mathbf{x}}\text{ (untargeted)},
\end{align}
where $\mathbf{\ddot x}$ and $\mathbf{\ddot x'}$ are the $2nd$-order derivatives of $\mathbf{x}$ and $\mathbf{x}'$, $w$ is a weight. Matching the $2nd$-order derivatives is proven to be important for visual imperceptibility in adversarial attack \cite{wang_understanding_2021}. $L$ is the Euclidean distance. $B_i$ and $B'_i$ are the $i$-th bone's lengths of the attacked and adversarial motion respectively. When the bone lengths change from frame to frame in the original data, we impose the bone-length constraint on each frame. $\theta'_j$ is the $j$-th joint angle in every pose of $\mathbf{x'}$ and subject to joint limits bounded by $\theta_j^{\min}$ and $\theta_j^{\max}$. Essentially, the optimization above seeks an adversarial sample that is: (1) close to the perturbed motion $\Tilde{\mathbf{x}}$ in terms of the Euclidean distance, (2) matching the motion dynamics to the original motion $\mathbf{x}$ and (3) on the manifold.

Equation~\ref{eq:gmw_Euclidean} is difficult to solve, especially to satisfy both the bone length and joint limit constraints in the joint position space \cite{zheng_towards_2020}. We therefore solve Equation~\ref{eq:gmw_Euclidean} in two steps. First, we solve it without any constraints by Inverse Kinematics in the joint angle space, which automatically preserves the bone lengths. Next, Equation~\ref{eq:gmw_Euclidean} is solved in the joint angle space:
\begin{align}
\label{eq:gmw_angles}
    & \min_{\mathbf{\theta'}}\ \ L(\widetilde{\mathbf{\theta}}, \mathbf{\theta'})  + w L(\mathbf{\ddot \theta}, \mathbf{\ddot \theta'})\ \ \text{s.t.}\ \ \theta_j^{\min} \le \mathbf{\theta'_j} \le \theta_j^{\max}, \nonumber \\
    & C_{\mathbf{x'}} = c\text{ (targeted)} \text{ or } C_{\mathbf{x'}} \ne C_{\mathbf{x}}\text{ (untargeted)}.
\end{align}
Note that the objective function in Equation~\ref{eq:gmw_angles} is designed to match the joint angles and the joint angular acceleration. We use a primal-dual interior-point method \cite{hedengren2014nonlinear} to solve Equation~\ref{eq:gmw_angles}. After solving for $\theta'$, the joint positions of the adversarial motion are computed using Forward Kinematics. Please refer to the supplementary materials for details of mathematical deduction.

\subsection{Mixed On-manifold Adversarial Training}
The assumption of adversarial training is that adversarial samples can help regulate classification boundaries to resist attacks~\cite{cvpr/Stutz/onm}. A common adversarial training (AT) strategy ~\cite{DBLP:conf/iclr/MadryMSTV18} is:
\begin{equation}
\label{eq:AT}
\min_{\theta}\mathbb{E}_{(\mathbf{x},y)\sim{D}} \ [\max_{\bm{\delta} \in {B(\epsilon)}} \mathcal{L}(\mathbf{x}+\bm{\delta},\bm{\theta},y)]
\end{equation}
where $D$ is the distribution over data pairs of $\mathbf{x} \in{\mathbb{R}^d}$ and label $y$. $\bm{\theta} \in{\mathbb{R}^p}$ is the model parameters. $B(\epsilon)=\{\bm{\delta} \ | \ \|\bm{\delta}\|_{\infty} \leq \epsilon $ \} is the perturbation set. $\mathcal{L}$ is a loss function, e.g. cross-entropy. During training, the perturbation $\delta$ is drawn from a prior e.g. Gaussian and uniform distribution, or some adversarial attack method, and restricted within the $\epsilon$-ball $B(\epsilon)$. 

One issue in Equation~\ref{eq:AT} is that there is an underlying assumption of the structural simplicity of the adversarial sample distribution in $B(\epsilon)$, which enables the usage of well-defined priors (e.g. Gaussians). However, we argue this assumption is overly simplified. The structure of the adversarial sample distribution can be arbitrarily complex. Consequently, when drawing perturbations from a prior, a conservative prior (e.g. Gaussians with too small variances) cannot resist attacks, while an aggressive one (e.g. Gaussians with too large variances) can be detrimental to the accuracy. On the other hand, drawing perturbations via attack leads to a more guided approximation of the adversarial sample distribution, but it also ignores the different importance across different adversarial samples. As a result, existing AT methods always need to compromise between accuracy and robustness~\cite{icml/Zhang/TRADES}.

Unlike existing methods, we explore a finer structure depicted by two distributions in $B(\epsilon)$: the distributions of on-manifold and off-manifold adversarial samples. We first assume the distribution of adversarial samples is different from the clean data~\cite{stutz_Disentangling_2019}. Next, we further make a more \textit{fine-grained} assumption: the distribution of on-manifold adversarial samples is different from that of the off-manifold adversarial samples. We expect the fine-grained distribution modeling to be able to eliminate the trade-off between accuracy and robustness, which remains unsolved currently. This is because on-manifold samples should be directly useful in simultaneously improving the accuracy and robustness, while the off-manifold samples are more aggressive and hence helpful in improving the robustness.

To this end, we propose a mixed manifold-based adversarial training (MMAT) which optimizes a hybrid loss consisting of a standard classification loss ($\mathcal{L}_{c}$), an on-manifold robustness loss ($\mathcal{L}_{on}$) and an off-manifold robustness loss ($\mathcal{L}_{off}$) term:
\begin{equation}
\label{eq:MMAT}
\min_{\theta}\mathbb{E}_{(\mathbf{x},y)\sim{D}} \ [\mu_{c}\mathcal{L}_{c} + \mu_{on}\mathcal{L}_{on} + \mu_{off} \mathcal{L}_{off}]
\end{equation}
where $\mu_{c}$, $\mu_{on}$ and $\mu_{off}$ are weights, $\mu_{c} = 1 - \mu_{on} - \mu_{off}$. The losses are:
\begin{align}
\label{eq:loss_c}
    &\mathcal{L}_{c} = \mathcal{L}(\mathbf{x},\bm{\theta},y))   \\
\label{eq:loss_on}
    &\mathcal{L}_{on} = \mathcal{L}(\mathbf{x}'_{on},\bm{\theta},y))\ \text{s.t.} \ \mathbf{x}'_{on} \in [0, 1]^{m\times n}, \ \mathbf{x}'_{on} \in \mathcal{M} \\
\label{eq:loss_off}
    &\mathcal{L}_{off} = \max\mathcal{L}(\mathbf{x}'_{off},\bm{\theta},y) \ \text{s.t.} \ \mathbf{x}'_{off} \in [0, 1]^{m\times n}
\end{align}

\noindent\textbf{Adversary Sampling:}
During optimization, we need to sample $\mathbf{x}'_{on}$ and $\mathbf{x}'_{off}$ as they cannot be described in any analytical form. We propose a black-box and a white-box sampling strategy: BASAR and SMART~\cite{wang_understanding_2021} with MMAT, named BASAR-MMAT and SMART-MMAT respectively. In BASAR-MMAT, $\mathbf{x}'_{on}$ and $\mathbf{x}'_{off}$ are generated by BASAR with/without manifold projection (BASAR-NoMP). In SMART-MMAT, $\mathbf{x}'_{on}$ and $\mathbf{x}'_{off}$ are generated by SMART~\cite{wang_understanding_2021} with/without perception loss.

\section{Attack Experiments}
\subsection{Settings}
To evaluate the proposed method, we conduct experiments on five state-of-the-art target models: ST-GCN~\cite{Yan:2018:STG}, MS-G3D~\cite{Liu_2020_CVPR}, SGN~\cite{Zhang_2020_CVPR_1}, CTR-GCN~\cite{ctrgcn} and FR-HEAD~\cite{frhead}. We choose not only the most popular benchmark datasets such as HDM05~\cite{cg-2007-2} and NTU 60~\cite{shahroudy2016ntu}, but also the challenging dataset UAV-Human~\cite{Li2021UAVHuman}, Kinetics-400~\cite{kay2017kinetics} and Skeletics-152~\cite{skeletics152}. UAV-Human~\cite{Li2021UAVHuman} was collected by an unmanned aerial vehicle and hence has low visual quality. Kinetics-400~\cite{kay2017kinetics} is a large and highly noisy dataset taken from different YouTube Videos. Similarly, Skeletics-152 is a 3D pose-annotated subset of videos sourced from much larger Kinetics-700 datasets~\cite{kinetics700}. The details of these databases are in the supplemental document. The experiments are conducted on a Xeon Silver 4216 CPU and an NVIDIA GTX 2080Ti GPU. The query time and number of queries are shown in the supplemental document.

\subsection{Evaluation Metrics}
\label{subsec:Metrics}
We employ the success rate as for evaluation. In addition, to further numerically evaluate the quality of the adversarial samples, we also define metrics between the original motion $\mathbf{x}$ and its adversarial sample $\mathbf{x}'$, including the averaged joint position deviation $l =  \frac{1}{n N} \sum_{j=0}^{N} \| \mathbf{x}^{(j)} - \mathbf{x}^{'(j)}\|_2$, the averaged joint acceleration deviation $\Delta a = \frac{1}{n O N} \sum_{j=0}^{N}\| \mathbf{\ddot x}^{(j)} - \mathbf{\ddot x}^{'(j)}\|_2$, the averaged joint angular acceleration deviation $\Delta \alpha =\frac{1}{n O N} \sum_{j=0}^{N}\| \mathbf{\ddot \theta}^{(j)} - \mathbf{\ddot \theta}^{'(j)}\|_2$, and the averaged bone-length deviation percentage $\Delta B/B = \frac{\sum_{j=0}^{N}\sum_{i=0}^{T}((B_{i}^{(j)}-B_{i}^{'(j)}) / B_i^{(j)})}{TN}$,
% \begin{align}
% \label{eq:metrics}
% l =  \frac{1}{n N} \sum_{j=0}^{N} \| \mathbf{x}^{(j)} - \mathbf{x}^{'(j)}\|_2   \nonumber \\ 
% \Delta a = \frac{1}{n O N} \sum_{j=0}^{N}\| \mathbf{\ddot x}^{(j)} - \mathbf{\ddot x}^{'(j)}\|_2  \nonumber \\
% \Delta \alpha =\frac{1}{n O N} \sum_{j=0}^{N}\| \mathbf{\ddot \theta}^{(j)} - \mathbf{\ddot \theta}^{'(j)}\|_2  \nonumber\\  
% \Delta B/B = \frac{\sum_{j=0}^{N}\sum_{i=0}^{T}((B_{i}^{(j)}-B_{i}^{'(j)}) / B_i^{(j)})}{TN} 
% \end{align}
where $N$ is the number of adversarial samples. $O$ and $T$ are the total number of joints and bones in a skeleton. $n$ is the number of poses in a motion. We also investigate the percentage of on-manifold (OM) adversarial motions. An attack sample is regarded as on-manifold if all its poses respect the bone-length and joint limit constraints. Finally, since Kinetics-400, UAV-Human and Skeletics-152 have missing joints, it is impossible to attack it in the joint angle space. So we only attack it in the joint position space. Consequently, $\Delta\alpha$ and OM cannot be computed on Kinetics-400, UAV-Human and Skeletics-152.

\subsection{Attack Evaluation}
To initialize for untargeted attack, we randomly sample a motion $\mathbf{x}'$ for a target motion $\mathbf{x}$ where $c_{\mathbf{x}'} \ne c_\mathbf{x}$. For Kinetics-400, UAV-Human and Skeletics-152, the maximum number of iterations is 1000, and it is set to 500 and 2000 for HDM05 and NTU 60 respectively. The results are shown in Table~\ref{tab:untargeted}. Note that BASAR achieves 100\% success in all tasks. Here we also conduct ablation studies (MP/No MP) to show the effects of the manifold projection. First, the universal successes across all datasets and models demonstrate the effectiveness of BASAR. The manifold projection directly affects the OM results. BASAR can generate as high as 99.55\% on-manifold adversarial samples. As shown in the perceptual study later, the on-manifold samples are very hard to be distinguished from the original motions even under very harsh visual comparisons. 
For targeted attack, the maximum iterations are set to 1000, 2000, 3000 on HDM05, Kinetics-400 and NTU 60. To initiate a targeted attack on $\mathbf{x}$, we randomly select a $\mathbf{x}'$ where $C_{\mathbf{x}'}$ = $c$ and $c$ is the targeted class. The results are shown in Table~\ref{tab:untargeted}. All attacks achieve 100\% success. The targeted attack is more challenging than the untargeted attack \cite{wang_understanding_2021}, because the randomly selected label often has completely different semantic meanings from the original one. Attacking an `eating' motion to `drinking' is much easier than to `running'. This is why the targeted attack, in general, has worse results than untargeted attack under every metric. Even under such harsh settings, BASAR can still produce as high as 56.98\% on-manifold adversarial samples. The performance variation across models is consistent with the untargeted attack.

\begin{table*}[!htb]
\caption{Untargeted attack(left) and targeted attack(right) on HDM05 (top), NTU (middle) and Kinetics-400 (bottom). All attacks have a 100\% success rate. $l$: averaged joint position deviation; $\Delta$a: averaged joint acceleration deviation; $\Delta\alpha$: averaged joint angular acceleration deviation; $\Delta$B/B: averaged bone-length deviation percentage; on-manifold sample percentage (OM). MP means Manifold Projection.}
\label{tab:untargeted}
\centering
\resizebox{1\linewidth}{!}{
\setlength{\tabcolsep}{1.2mm}{
\begin{tabular}{c|c|ccccc|ccccc}

\hline
& & \multicolumn{5}{c|}{Untargeted Attack} & \multicolumn{5}{c}{Targeted Attack}\\
Models&  & $l$$\downarrow$ & $\Delta a$$\downarrow$ & $\Delta\alpha\downarrow$ & $\Delta B/B$$\downarrow$ & OM$\uparrow$ & $l$$\downarrow$ & $\Delta a$$\downarrow$ & $\Delta\alpha\downarrow$ & $\Delta B/B$$\downarrow$ & OM$\uparrow$\\
\hline
\multirow{2}{*}{ST-GCN} & MP & 0.13  & 0.05 & 0.11 & 0.00\% & 99.55\% & 4.97  & 0.10 & 0.65  & 3.44\% & 56.98\% \\
& No MP & 0.10 & 0.04 & 0.34 & 0.66\% & 0.00\% & 6.25  & 0.09 & 0.92  & 5.85\% & 0.00\% \\
\hline
\multirow{2}{*}{MS-G3D} & MP & 0.76  & 0.12 & 0.49  & 1.78\% & 0.13\% & 4.34  & 0.12 & 0.71  & 4.51\% & 1.64\%\\
& No MP & 0.70  & 0.09 & 0.82  & 1.81\% & 0.00\% & 4.35  & 0.11 & 1.01  & 5.08\% & 0.00\%\\
\hline
\multirow{2}{*}{SGN} & MP & 11.53 & 1.92 & 6.70 & 9.60\% & 60.52\% & 16.31 & 1.28 & 6.97 & 12.29\% & 20.96\%\\
& No MP & 7.93 & 2.00 & 14.36 & 39.64\% & 0.00\%  &16.13 & 1.63 & 13.28 & 29.86\% & 0.00\%\\ 
\hline\hline
\multirow{2}{*}{ST-GCN} & MP & 0.08 & 0.02 & 0.07 & 4.82\% & 4.68\%  & 0.37 & 0.03 & 0.25 & 9.73\% & 0.63\%\\
& No MP & 0.10 & 0.02 & 0.09 & 5.57\% & 1.82\%  & 0.38 & 0.04 & 0.16 & 11.55\% & 0.16\%\\ 
\hline
\multirow{2}{*}{MS-G3D} & MP & 0.08 & 0.03 & 0.12 & 8.14\% & 0.86\%  & 0.36 & 0.05 & 0.24 & 15.43\% & 0.00\%\\
& No MP & 0.12 & 0.03 & 0.17 & 10.02\% & 0.57\% & 0.40 & 0.06 & 0.27 & 17.72\% & 0.00\% \\ 
\hline
\multirow{2}{*}{SGN} & MP & 0.28 & 0.08 & 0.21 & 11.11\% & 28.95\% & 1.28 & 0.09 & 0.38 & 28.24\% & 2.63\%\\
& No MP & 0.30 & 0.10 & 0.42 & 28.00\% & 4.55\% & 1.35 & 0.10 & 0.53 & 39.43\% & 0.00\%  \\ 
\hline\hline
\multirow{2}{*}{ST-GCN} & MP & 0.05 & 0.0057 & n/a & 2.54\% & n/a  & 0.63 & 0.03 & n/a & 29.10\% & n/a \\
& No MP & 0.07 & 0.0062 & n/a & 3.53\% & n/a   & 0.67 & 0.03 & n/a & 31.48\% & n/a \\ 
\hline
\multirow{2}{*}{MS-G3D} & MP & 0.10 & 0.011 & n/a & 5.16\% & n/a  & 0.56 & 0.05 & n/a & 27.26\% & n/a\\
& No MP & 0.10 & 0.012 & n/a & 5.69\% & n/a  & 0.57 & 0.07 & n/a & 28.35\% & n/a \\ 
\hline
\multirow{2}{*}{SGN} & MP & 0.12 & 0.020 & n/a & 4.23\% & n/a  & 1.51 & 0.18 & n/a & 68.45\% & n/a \\
& No MP & 0.13 & 0.022 & n/a & 6.93\% & n/a  & 1.54 & 0.19 & n/a & 72.09\% & n/a \\ 
\hline
\end{tabular}}}
\end{table*}
%we randomly select the same number of motions from each dataset as in untargeted attack, 
\paragraph{Attack on latest classifiers and datasets} CTR-GCN~\cite{ctrgcn} and FR-HEAD~\cite{frhead} are the recently proposed classifiers with more robust skeleton representations. We hence investigate the effectiveness of BASAR on the two challenging target models. As shown in Table~\ref{tab:latest}, BASAR can still generate manifold adversarial samples with a high probability even when the target models have robust representations. We also test BASAR on the latest wild human motion dataset UAV-Human~\cite{Li2021UAVHuman} and Skeletics-152~\cite{skeletics152}, and the results are reported in Table~\ref{tab:uav}. The data quality of UAV-Human and Skeletics-152 is similar to Kinetics-400 and so is the attack performance. Overall, BASAR with manifold projection can improve the attack quality via reducing the $l$, $\Delta a$ and $\Delta B/B$.

% % We speculate that this is due to the low data quality and classification accuracy (30.25\% on ST-GCN). Somewhat counter-intuitively, bad classification boundaries make it hard to attack. Nevertheless, BASAR with manifold projection can still generate natural-looking adversarial samples.

\begin{table}[!htb]
\caption{(Untargeted) Attack performance on latest classifiers. All attacks have a 100\% success rate.}
\label{tab:latest}
\centering
\resizebox{0.9\linewidth}{!}{
\setlength{\tabcolsep}{1.2mm}{
\begin{tabular}{c|c|ccccc|ccccc}
\hline
\multirow{2}{*}{Models} &\multirow{2}{*}{}  & \multicolumn{5}{c|}{HDM05} & \multicolumn{5}{c}{NTU 60} \\
&  & $l$$\downarrow$ & $\Delta a$$\downarrow$ & $\Delta\alpha\downarrow$ & $\Delta B/B$$\downarrow$ & OM$\uparrow$ & $l$$\downarrow$ & $\Delta a$$\downarrow$ & $\Delta\alpha\downarrow$ & $\Delta B/B$$\downarrow$ & OM$\uparrow$ \\
\hline
\multirow{2}{*}{CTR-GCN} & MP & 0.67  & 0.14 & 0.31 & 0.80\% & 45.0\% & 0.05 & 0.02 & 0.03 & 6.50\% & 2.2\%\\
& No MP & 0.63 & 0.13 & 1.00 & 2.18\% & 1.4\% & 0.07 & 0.02 & 0.04 & 7.14\% & 0.8\%\\
\hline
\multirow{2}{*}{FR-HEAD} & MP & 0.15  & 0.06 & 0.07  & 0\% & 93.2\% & 0.04 & 0.013 & 0.06 & 4.13\% & 10.8\%\\
& No MP & 0.16  & 0.05 & 0.42  & 0.76\% & 5.9\% & 0.06 & 0.014 & 0.07 & 5.14\% & 7.9\%\\
\hline
\end{tabular}}}
\end{table}

% % \begin{table}[tb]
% % \centering
% % \caption{ST-GCN Attack on UAV-Human. SR means attack success rate.}
% % \label{tab:uav}
% % \setlength{\tabcolsep}{1.2mm}{
% % \begin{tabular}{c|c|cccc}
% % \hline
% % Attack Type&  & $l$$\downarrow$ & $\Delta a$$\downarrow$ &  $\Delta B/B$$\downarrow$ & SR \\
% % \hline
% % \multirow{2}{*}{Untargeted} & No MP & 21.83  & 11.41   & 8.05\% & 100\% \\
% % & MP & 17.68  & 9.30  & 6.81\% & 100\%  \\ 
% % \hline\hline
% % \multirow{2}{*}{Targeted} & No MP & 69.51 & 32.24  & 28.77\% & 63.91\%  \\
% % & MP & 60.07 & 30.59 & 25.10\% & 63.07\%  \\ 
% % \hline
% % \end{tabular}}
% % \end{table}

\begin{table}[!htb]
\centering
\caption{(Untargeted) Attack performance on the wild human motion dataset with ST-GCN. All attacks have a 100\% success rate.}
\label{tab:uav}
\resizebox{0.6\linewidth}{!}{
\begin{tabular}{c|ccc|ccc}
\hline
\multirow{2}{*}{ST-GCN}      & \multicolumn{3}{c|}{UAV} & \multicolumn{3}{c}{Skeletics-152} \\
    & $l$$\downarrow$    & $\Delta a$$\downarrow$    & $\Delta B/B$$\downarrow$    & $l$$\downarrow$       & $\Delta a$$\downarrow$       & $\Delta B/B$$\downarrow$       \\\hline
MP   & 17.68     &  9.30      & 6.81\%        & 0.095   & 0.019     &  3.73\%          \\
NO MP & 21.83    & 11.41    & 8.05\%       & 0.098   &0.019         & 3.88\%         \\\hline
\end{tabular}}
\end{table}

\subsection{Perceptual Studies} Numerical evaluation alone is not sufficient to evaluate the imperceptibility of adversarial attack on skeletal motions, because they cannot accurately indicate whether the attack is perceptible to humans \cite{wang_understanding_2021}. We, therefore, conduct rigorous perceptual studies to evaluate the imperceptibility of BASAR. Therefore, we design a new perceptual study protocol that includes three perception metrics: Deceitfulness, Naturalness and Indistinguishability. Deceitfulness is to test whether BASAR visually changes the semantics of the motion. Naturalness aims to test whether on-manifold adversarial samples look more natural than off-manifold adversarial samples. Indistinguishability is the strictest test to see whether adversarial samples by BASAR can survive a side-by-side scrutiny. The details of the perceptual study is reported in \textbf{Supplemental Document}. %The results show that BASAR is more deceitful than off-manifold adversarial examples. More importantly, BASAR can even survive under the strictest side-by-side scrutiny.

The success rate of \textit{Deceitfulness} is 79.64\%. Next, 85\% of the on-manifold adersarial samples look more natural than off-manifold samples. This is understandable as manifold projection not only makes sure the poses are on the manifold, but also enforces the similarity of the dynamics between the attacked and original motion. Finally, the results of \textit{Indistinguishability} are 89.90\% on average. BASAR even outperforms the white-box attack (80.83\%) in \cite{wang_understanding_2021}. BASAR successfully fools the users under the strictest side-by-side scrutiny. The complete perceptual evaluation is reported in \textbf{Supplemental Document}

\subsection{Comparison}
Since BASAR is the very first black-box adversarial attack method on skeletal motions, there is no baseline for comparison. So we employ methods that are closet to our approach as baselines. Although SMART~\cite{wang_understanding_2021} is designed for white-box attack, it can also be used as transfer-based attack via requiring surrogate models. We hence employ it as a baseline and choose HRNN \cite{Du:2015:HRN} and 2SA-GCN \cite{Shi_2019_CVPR} as the surrogate models. The second baseline is MTS \cite{karim_adversarial_2020} which is a black-box method but only on general time-series. It is the most similar method to BASAR but does not model the data manifold. Another baseline is BA \cite{drendel_decisionaa_2018}, a decision-boundary based attack designed for images. We choose HDM05, NTU 60 and Skeletics 152 for comparisons. Since MTS is not designed for untargeted attack, we only compare BASAR with it on the targeted attack. 
%For each comparison, we randomly select 1200 and 490 motions from NTU and HDM05. 
%For SMART, we randomly select 1200 and 490 results on each attack on NTU and HDM05 dataset.\HW(how many motions?). In addition, since MTS is not designed for untargeted attack, we attack every motion with a specific label \HW(how did you choose this label?) in validation set \HW(again, how many motions are in this validation set?) of NTU and HDM05. To compare with BA, we attack the same target motions (\HW(don't get it. Why not attack the same target motions using SMART and MTS?)) with BASAR.
%The first is SMART \cite{wang_understanding_2021} which is a white-box approach. Although it can deliver a black-box attack, it needs a surrogate model. According to their work, we choose HRNN \cite{Du:2015:HRN} and 2SAGCN \cite{Shi_2019_CVPR} as the surrogate models.
\begin{table}[tb]
\caption{Attack success rate comparison with baseline methods. S-152 means the Skeletics-152 dataset.}
\label{tab:comparison1}
\centering
\resizebox{0.8\linewidth}{!}{
\begin{tabular}{c|c|ccc|ccc}
\hline
\multirow{2}{*}{Models} & \multirow{2}{*}{Attacked Method} & \multicolumn{3}{c|}{Untargeted Attack} & \multicolumn{3}{c}{Targeted Attack}  \\
                        &           & HDM05       & NTU  &  S-152     & HDM05       & NTU   &  S-152         \\ \hline
\multirow{4}{*}{ST-GCN}  & BASAR         & 100\%      & 100\%   &100\%       & 100\%       & 100\%   & 100\% \\ 
                        & MTS           & n/a        & n/a      & n/a     & 3.3\%      & 12.0\%      &4.7\%\\
                        &SMART(HRNN)    & 66.9\%    & 89.3\%  & 33.6\%     & 3.2\%      & 2.3\%      &2.2\%\\
                        &SMART(2SAGCN)  & 86.1\%    & 12.9\%  & 14.1\%    & 2.3\%      & 0.2\%       &1.1\%\\
                        \hline\hline
\multirow{4}{*}{MS-G3D}  & BASAR         & 100\%      & 100\%   & 100\%       & 100\%       & 100\%   & 100\% \\ 
                        & MTS           & n/a        & n/a      &n/a      & 2.2\%      & 12.9\%   & 3.6\% \\
                        &SMART(HRNN)    & 86.9\%    & 99.3\%  &51.1\%      & 30.3\%     & 1.2\%  &1.1\%  \\
                        &SMART(2SAGCN)  & 88.7\%    & 3.1\%   & 2.3\%      & 2.5\%      & 0.00\%  &0.6\%  \\
                        \hline\hline
\multirow{4}{*}{SGN}    & BASAR         & 100\%      & 100\%  &100\%         & 100\%       & 100\%   &100\%  \\ 
                        & MTS           & n/a        & n/a    &n/a           & 2.91\%      & 0.00\%  & 1.4\%  \\
                        &SMART(HRNN)    & 89.25\%    & 98.25\%  &20.3\%      & 29.69\%     & 1.42\%  & 2.1\%  \\
                        &SMART(2SAGCN)  & 0.41\%     & 97.75\%  &10.6\%      & 3.28\%      & 1.83\%  & 1.3\%  \\
                        \hline
\end{tabular}}
\end{table}

Table~\ref{tab:comparison1} lists the success rates of all methods. BASAR performs the best and often by big margins. In the targeted attack, the highest attack success rate among the baseline methods is merely 30.3\% on HDM05, 12.9\% on NTU and 4.7\% on Skeletics-152 while BASAR achieves 100\%. In the untargeted attack, the baseline methods achieve higher performances but still worse than BASAR. SMART achieves as high as 99.33\% on NTU/MS-G3D. However, its performance is not reliable as it highly depends on the chosen surrogate model, which is consistent with \cite{wang_understanding_2021}. In addition, we further look into the results and find that SMART's results are inconsistent. When the attack is transferred, the result labels are often different from the labels obtained during the attack. 

We find that BA can also achieve 100\% success. However, BA is designed to attack image data and does not consider the data manifold. {We therefore compare detailed metrics and show the results in Table~\ref{tab:comparison2}. BA is in general worse than BASAR under every metric. The worst is the bone-length constraint violation. Visually, the skeletal structure cannot be observed at all. This happens for both the untargeted and the targeted attack across all datasets and models. This is understandable because BA does not consider the data manifold, while BASAR assumes that in-manifold perturbations provide better visual quality. One qualitative comparison with BA can be found in Figure~\ref{fig:comparsion}.

\begin{table}[!htb]
\caption{Boundary Attack (BA) on HDM05 (left), NTU (right). UA/TA refers Untargeted/Targeted Attack.} %All achieved 100\% successful rate. We show the the averaged joint position deviation ($l$), averaged joint acceleration deviation ($\Delta$a), averaged joint angular acceleration deviation ($\Delta\alpha$), averaged bone-length deviation percentage ($\Delta$B/B), on-manifold sample percentage (OM). UA/TA refers Untargeted/Targeted attack.}
\label{tab:comparison2}
\centering
\resizebox{0.8\linewidth}{!}{
\setlength{\tabcolsep}{1.2mm}{
\begin{tabular}{c|c|ccccc|ccccc}
\hline
Models&  & \textit{l}$\downarrow$ & $\Delta$a$\downarrow$ & $\Delta\alpha\downarrow$ & $\Delta$B/B$\downarrow$ & OM$\uparrow$ & \textit{l}$\downarrow$ & $\Delta$a$\downarrow$ & $\Delta\alpha\downarrow$ & $\Delta$B/B$\downarrow$ & OM$\uparrow$\\
\hline
\multirow{2}{*}{ST-GCN} & UA & 1.44 & 0.65& 4.74 & 10.60\% & 0.00\% & 1.04 & 0.47 & 1.97 & 235.10\% & 0.00\%\\
& TA & 8.83 & 0.17 & 1.60 & 8.56\% & 0.00\% & 1.04 & 0.47 & 1.97 & 235.10\% & 0.00\%\\
\hline
\multirow{2}{*}{MS-G3D} & UA & 1.17  & 0.36 & 2.81  & 6.00\% & 0.00\% & 1.24 & 1.73 & 2.38 & 911.7\% & 0.00\%\\
& TA & 7.93  & 0.10 & 1.07  & 7.49\% & 0.00\% & 0.27 &0.07  & 0.34 & 25.72\% & 0.00\% \\
\hline
\multirow{2}{*}{SGN} & UA & 13.35 & 3.45 & 21.96 & 75.11\% & 0.00\% & 0.22 & 0.28 & 1.26  & 125.57\% & 3.60\%\\
& TA & 15.40 & 1.52 & 11.77 & 29.45\% & 0.00\% & 0.42 & 0.15 & 0.66  & 65.31\% & 0.17\% \\  
\hline
\end{tabular}}}
\end{table}

% \begin{table}[!htb]
% \caption{Boundary Attack (BA) on HDM05 (left), NTU (right). UA/TA refers Untargeted/Targeted Attack.} %All achieved 100\% successful rate. We show the the averaged joint position deviation ($l$), averaged joint acceleration deviation ($\Delta$a), averaged joint angular acceleration deviation ($\Delta\alpha$), averaged bone-length deviation percentage ($\Delta$B/B), on-manifold sample percentage (OM). UA/TA refers Untargeted/Targeted attack.}
% \label{tab:comparison2}
% \centering
% \resizebox{0.8\linewidth}{!}{
% \setlength{\tabcolsep}{1.2mm}{
% \begin{tabular}{c|c|ccccc|ccccc}
% \hline
% Models&  & \textit{l}$\downarrow$ & $\Delta$a$\downarrow$ & $\Delta\alpha\downarrow$ & $\Delta$B/B$\downarrow$ & OM$\uparrow$ & \textit{l}$\downarrow$ & $\Delta$a$\downarrow$ & $\Delta\alpha\downarrow$ & $\Delta$B/B$\downarrow$ & OM$\uparrow$\\
% \hline
% \multirow{2}{*}{ST-GCN} & UA & 1.44 & 0.65& 4.74 & 10.60\% & 0.00\% & 1.04 & 0.47 & 1.97 & 235.10\% & 0.00\%\\
% & TA & 8.83 & 0.17 & 1.60 & 8.56\% & 0.00\% & 1.04 & 0.47 & 1.97 & 235.10\% & 0.00\%\\
% \hline
% \multirow{2}{*}{MS-G3D} & UA & 1.17  & 0.36 & 2.81  & 6.00\% & 0.00\% & 1.24 & 1.73 & 2.38 & 911.7\% & 0.00\%\\
% & TA & 7.93  & 0.10 & 1.07  & 7.49\% & 0.00\% & 0.27 &0.07  & 0.34 & 25.72\% & 0.00\% \\
% \hline
% \multirow{2}{*}{SGN} & UA & 13.35 & 3.45 & 21.96 & 75.11\% & 0.00\% & 0.22 & 0.28 & 1.26  & 125.57\% & 3.60\%\\
% & TA & 15.40 & 1.52 & 11.77 & 29.45\% & 0.00\% & 0.42 & 0.15 & 0.66  & 65.31\% & 0.17\% \\  
% \hline
% \end{tabular}}}
% \end{table}

\begin{figure}[!htb]
    \centering
    \includegraphics[width=0.8\linewidth]{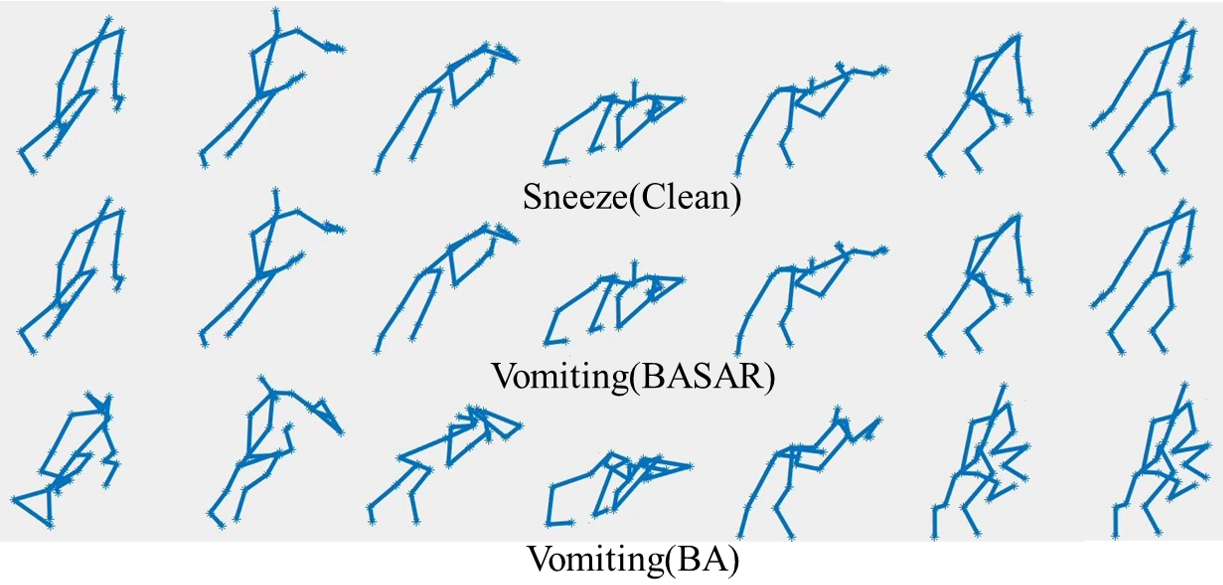}
    \caption{The visual comparison with BA. The first row is the clean motion labeled as `Squeeze'. The second row is the adversarial motion generated by BASAR and misclassified as `Vomiting'. The third row is the adversarial motion generated by BA and misclassified as `Vomiting'.}
    \label{fig:comparsion}
\end{figure}

\begin{table}[!htb]
\caption{The results of Rays and Rays with manifold projection(Rays-MP) on HDM05(left) and NTU 60(right).  $l_{\infty}$ means the $l_{\infty}$ norm distance between adversarial examples and original examples. SR means attack success rate. We show the best performance with bold.}
    \centering
    \label{tab:rays}
    \setlength{\tabcolsep}{1 mm}{
\resizebox{1\linewidth}{!}{
\begin{tabular}{c|c|ccccc|c|ccccc|c}
\hline
\multirow{2}{*}{Models} &\multirow{2}{*}{Attacks} & \multicolumn{5}{c|}{Numerical Evaluation}                     & \multirow{2}{*}{$l_{\infty}$(0.5)} & \multicolumn{5}{c|}{Numerical Evaluation}                     & \multirow{2}{*}{$l_{\infty}$(0.05)}\\
& &\textit{l}$\downarrow$ & $\Delta$a$\downarrow$ & $\Delta\alpha\downarrow$ & $\Delta$B/B$\downarrow$ & OM$\uparrow$   &   & \textit{l}$\downarrow$ & $\Delta$a$\downarrow$ & $\Delta\alpha\downarrow$ & $\Delta$B/B$\downarrow$ & OM$\uparrow$   & \\
\hline
\multirow{2}{*}{ST-GCN} & Rays& 0.08 & 0.015 & 0.19 & 0.02\% & 27.2\%   & 100\% & 0.061 & 0.028 & 0.019 & 4.6\% & 2.2\%  & 99.9\%\\

& Rays-MP& \textbf{0.07} & \textbf{0.013} & \textbf{0.13} & \textbf{0.01\%} & \textbf{56.7\%}  & 100\% &\textbf{0.056} & \textbf{0.026} & \textbf{0.018} & \textbf{4.2\%} & \textbf{4.8\%}  & \textbf{100\%}\\
\hline
\multirow{2}{*}{MS-G3D} & Rays& 1.67 & 0.07 & 0.28 & 0.11\% & 0\% & 95.2\% & 0.069 & 0.0154 & 0.050 & 2.86\% & 1.2\% & 98.8\%\\

& Rays-MP& \textbf{1.57} & \textbf{0.06} & \textbf{0.25} & \textbf{0.07\%} &\textbf{3.1\%} & \textbf{96.9\%} & \textbf{0.039} & \textbf{0.0146} & \textbf{0.047} & \textbf{2.61\%} &\textbf{3.8\%} &\textbf{100\%}\\
\hline
\multirow{2}{*}{SGN} & Rays& 1.35 & 0.17 & 0.2 & 0.1\% & 60.8\%  & 73.1\% & 0.19 & 0.003 & 0.0006 & 0.32\% & 95.9\%   & 95.9\%\\
& Rays-MP& \textbf{0.89} & \textbf{0.06} & \textbf{0.1} & \textbf{0.0\%} &\textbf{94.0\%} & \textbf{96.0\%} & \textbf{0.07} & \textbf{0.0003} & \textbf{0.0001} & \textbf{0.27\%} &\textbf{99.3\%} & \textbf{98.5\%}\\
                       
% \hline\hline
% \multirow{2}{*}{ST-GCN} & Rays& 0.061 & 0.028 & 0.019 & 4.6\% & 2.2\% & \multirow{2}{*}{0.05}        & 99.9\% \\

%                        & Rays-MP& \textbf{0.056} & \textbf{0.026} & \textbf{0.018} & \textbf{4.2\%} & \textbf{4.8\%} &  & \textbf{100\%} \\
% \hline
% \multirow{2}{*}{MS-G3D} & Rays& 0.069 & 0.0154 & 0.050 & 2.86\% & 1.2\% & \multirow{2}{*}{0.05}        & 98.8\% \\

%                        & Rays-MP& \textbf{0.039} & \textbf{0.0146} & \textbf{0.047} & \textbf{2.61\%} &\textbf{3.8\%} &         & \textbf{100\%} \\
% \hline
% \multirow{2}{*}{SGN} & Rays& 0.19 & 0.003 & 0.0006 & 0.32\% & 95.9\% & \multirow{2}{*}{0.05}        & 95.9\% \\

%                        & Rays-MP& \textbf{0.07} & \textbf{0.0003} & \textbf{0.0001} & \textbf{0.27\%} &\textbf{99.3\%} &         & \textbf{98.5\%} \\
\hline
\end{tabular}}
}
%\vspace{-0.1cm}
\end{table}
\subsubsection{Effectiveness of Manifold Projection}
The manifold projection is a general operation which can theoretically work with other attackers. To verify this, we adapt Rays~\cite{Rays}, a state-of-the-art decision-based attack for images, to attack motions by treating a motion sample as an image. Based on our experiments, Rays fails in targeted attack even with manifold projection, which is not surprising as it is not designed for attacking motions. Therefore, we only report the untargeted attack results. The model queries are the same as BASAR. As shown in Table~\ref{tab:rays}, Rays-MP with manifold projection can always improve the attack success rate and the attack quality via reducing the \textit{l}, $\Delta$a, $\Delta\alpha$ and $\Delta$B/B metrics. Moreover, Rays-MP can generate more natural adversarial motions.

\section{Defense Experiments}
\label{Sec:Def}
\subsection{Experiment Setup}
We evaluate MMAT on HDM05~\cite{cg-2007-2} using ST-GCN~\cite{Yan:2018:STG}, MS-G3D~\cite{Liu_2020_CVPR} and SGN~\cite{Zhang_2020_CVPR_1}. NTU 60~\cite{shahroudy2016ntu} and Kinetics-400~\cite{kay2017kinetics} are excluded for their extensive noises making it difficult to evaluate the effects of on-manifold samples in AT. We follow the original settings~\cite{Yan:2018:STG,Liu_2020_CVPR,Zhang_2020_CVPR_1} to train these networks. For BASAR-MMAT, we regard adversarial examples generated by BASAR as data augmentation. For SMART-MMAT, we use SMART-50 (SMART with 50 iterations) for training. After adversarial training, we attack the trained model with BASAR-NoMP with 500 iterations, as it generates more violent attacks than BASAR. We also employ SMART-200~\cite{wang_understanding_2021} and CIASA-200~\cite{liu_adversarial_2019} to test the classifier robustness under white-box attacks. Since there are three weights in our adversarial training loss, we conduct an ablation study to identify the optimal weights in different settings in~\textbf{Supplemental Document}.

\subsection{Robustness Evaluation}
\label{sec:compar}
We employ TRADES~\cite{icml/Zhang/TRADES} and MART~\cite{wang2020improving} as baselines, which are the state-of-the-art AT methods on images. Zheng et al.~\cite{zheng_towards_2020} use an adapted random smoothing (RS)~\cite{Cohen19_ICML} approach for defending skeleton-based HAR, so we use it as one baseline. The results are shown in Table \ref{tab:compar}. First, both BASAR-MMAT and SMART-MMAT can improve the robustness (Table~\ref{tab:compar}) and SMART-MMAT has slightly better overall performance. This is understandable since SMART-MMAT is white-box and computes more aggressive adversaries. However, BASAR-MMAT can achieve better accuracy than standard training and SMART-MMAT, which show BASAR-MMAT can potentially eliminate the accuracy-robustness trade-off. Next, SMART-MMAT is apparently more robust than RS and MART, and outperforms TRADES under most attack scenarios. More importantly, our method not only improves the robustness but also not compromise standard accuracy. The MMAT accuracy is within a small range(+3.76\%/-0.59\%) from that of standard training, in contrast to the noticeable accuracy decreased in other baseline methods. This is because the natural motion manifold is not considered in baseline methods, which means the generated adversarial samples are far away from the motion manifold, decreasing the standard accuracy. 

To further understand the reason, we plot the deviation distributions of on/off-manifold adversarial samples across three models. The deviation is computed based on the $l_{2}$ distance between each clean sample and its corresponding adversarial sample. Since the distributions are similar across the three models, we only shown the distribution on ST-GCN (Figure~\ref{fig:on_off_dist}). From the figure, we can see that there are long tails in both distributions. Further there is clearly more than one mode when combining both distributions. Random Smoothing~\cite{Cohen19_ICML} essentially expands the data distribution homogeneously and symmetrically by a fixed distance at every data point, while TRADES and MART draw perturbations via specific adversary. Their strategy is overly simplified and hence cannot capture the fine-grained structure of the adversarial distribution like MMAT. This shows the necessity of modeling the distributions of on/off manifold samples separately

%Figure~\ref{fig:on_off_dist} first confirms the validity of Gaussian Smoothing as there is a single mode in the on-manifold distribution near zero. A properly designed Gaussian noise could capture mass in this distribution. However, when we combine the two distributions in Figure~\ref{fig:on_off_dist}, it is far from single-modal. There are long tails in both distributions. Further there is clearly more than one mode when combining both distributions. Gaussian Smoothing essentially expands the data distribution homogeneously and symmetrically by a fixed distance at every data point ~\cite{Lecuyer_certified,Cohen19_ICML}. This strategy is overly simplified, not only making the performance sensitive to the manually tuned hyperparameters (e.g. the covariances), but also introducing the common trade-off between accuracy and robustness. This shows the necessity of modeling the distributions of on/off manifold samples separately and non-parametrically.

\begin{figure}[!htb]
    \centering
    \includegraphics[width=0.8\linewidth]{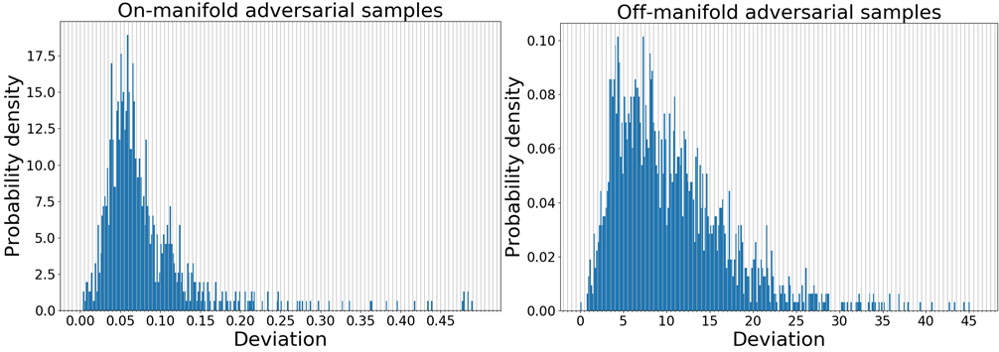}
    \caption{Deviation distributions of on/off-manifold adversarial samples (attacking ST-GCN) on HDM05 dataset.}
    \label{fig:on_off_dist}
\end{figure}

\begin{table}[!htb]
\caption{Comparing robustness with different defense methods. Top: ST-GCN, Middle: MS-G3D, Bottom: SGN. Acc is the classification accuracy, i.e. average accuracy on clean data. ST means standard training. MMAT means SMART-MMAT.}
\label{tab:compar}
\centering
\resizebox{1\linewidth}{!}{
\begin{tabular}{cc|ccc|cc|cc|c}
\hline
\multirow{2}{*}{Models} &\multirow{2}{*}{Methods}  &\multicolumn{3}{c|}{BASAR-NoMP} &\multicolumn{2}{c|}{SMART}  &\multicolumn{2}{c|}{CIASA} &\multirow{2}{*}{Acc}\\
&& $l\uparrow$ & $\Delta a\uparrow$ & $\Delta B/B\uparrow$ &@50 & @200  & @50 & @200 &\\\hline
\multirow{6}{*}{ST-GCN} &ST &0.1 &0.04 &0.66\% & 3.76\%  & 0.58\%  & 1.69\% &0.58\% & 87.2\% \\
&BASAR-MMAT &2.07 &0.67 &10.91\% &47.54\% &92.63\% &48.75\% &93.75\% &\textbf{91.2\%} \\
%GS($\sigma=0.01$) &0.61 &0.20 &4.13\% &86.16\% &99.11\% &86.61\% &99.33\% &\textbf{91.9\%}\\
% GS($\sigma=0.05$) &1.98 &0.68 &10.89\% &43.30\% &92.41\% &48.05\% &95.31\% &90.6\% \\
&RS &2.80 &1.16 &19.72\% &64.57\% &14.33\% &62.55\% &11.71\% &90.4\%\\ 
% GS($\sigma=0.5$) &9.86 &4.43 &94.93\% &5.36\% &31.47\% &6.77\% &40.63\% &89.7\% \\
&MART &\textbf{6.26} &2.14 &39.27\% &64.84\% &52.50\% &63.62\% &49.33\% &85.2\% \\
&TRADES &5.25 &1.72 &30.68\% &83.21\% &\textbf{65.83\%} &81.78\% &\textbf{63.54\%} &85.8\% \\
&\textbf{SMART-MMAT} &5.98 &\textbf{2.56} &\textbf{47.98\%} &\textbf{85.68\%}  & 62.33\%  & \textbf{85.31\%} &60.91\% & \textbf{91.0\%} \\
\hline\hline
\multirow{6}{*}{MS-G3D} &ST &0.70 &0.09 &1.81\%  & 6.59\%  & 1.47\%  & 6.10\% &4.19\% & 94.4\% \\
&BASAR-MMAT &4.53 &0.86 &15.86\% &77.23\% &92.41\% &70.76\% &91.13\% &\textbf{95.9\%} \\
%GS($\sigma=0.01$) &2.16 &0.67 &11.09\% &49.06\% &97.50\% &51.62\% &97.69\% &94.4\% \\ 
% GS($\sigma=0.05$) &4.91 &1.85 &32.70\% &8.75\% &63.74\% &7.92\% &65.00\% &94.5\% \\
&RS &6.90 &2.59 &48.08\% &90.35\% &39.56\% &90.57\% &57.44\% &93.4\% \\
% GS($\sigma=0.5$)  &13.60 &5.95 &132.30\% &2.29\% &17.50\% &2.29\% &17.50\% &93.0\% \\ 
&MART &8.71 &3.70 &77.26\% &88.09\% &81.84\% &87.70\% &80.27\% &90.6\% \\
&TRADES &\textbf{10.43} &\textbf{4.19} &\textbf{87.34\%} &88.83\% &84.33\% &88.63\% &82.90\% &90.0\% \\
&\textbf{SMART-MMAT} &10.30 &3.97 &79.07\% & \textbf{91.69\%}  & \textbf{84.79\%}  & \textbf{91.71\% }&\textbf{85.06\%}&\textbf{93.8\%} \\
\hline\hline
\multirow{6}{*}{SGN}&ST  &7.93 &2.00 &39.64\% & 63.29\%  & 22.56\%  & 69.15\%  &22.56\% & 94.1\% \\
%BASAR-MMAT &\textbf{14.85} &\textbf{3.76} &\textbf{80.26\%} &20.31\% &71.09\% &22.10\% &70.31\% &\textbf{94.7\%} \\
%GS($\sigma=0.01$) &6.13 &1.10 &19.10\% &32.29\% &90.63\% &35.04\% &91.52\% &\textbf{94.7\%} \\
% GS($\sigma=0.05$) &12.36 &2.90 &59.58\% &24.22\% &74.48\% &22.19\% &77.19\% &93.41\% \\
&RS &12.87 &3.07 &64.18\% &79.14\% &32.78\% &76.71\% &24.76\% &93.2\% \\
% GS($\sigma=0.5$)  &15.53 &4.23 &95.41\% &8.07\% &38.02\% &7.81\% &42.19\% &92.53\% \\
&MART &8.46 &2.04 &40.46\% &84.77\% &31.25\% &83.79\% &27.92\% &92.1\% \\
&TRADES &8.46 &2.04 &40.44\% &84.18\% &36.53\% &82.62\% &25.08\% &91.2\% \\
&\textbf{MMAT} &\textbf{13.06} &\textbf{3.46} &\textbf{65.69\%} &\textbf{87.36\%}  &\textbf{48.44\%}  & \textbf{87.00\%} &\textbf{45.15\%} & \textbf{93.9\%} \\
\hline\hline
\multirow{6}{*}{FR-HEAD}&ST  &0.16 &0.05 &0.76\% & 19.0\% & 4.9\%  & 14.3\%  &4.2\% & 93.4\% \\
%BASAR-MMAT &\textbf{14.85} &\textbf{3.76} &\textbf{80.26\%} &20.31\% &71.09\% &22.10\% &70.31\% &\textbf{94.7\%} \\
%GS($\sigma=0.01$) &6.13 &1.10 &19.10\% &32.29\% &90.63\% &35.04\% &91.52\% &\textbf{94.7\%} \\
% GS($\sigma=0.05$) &12.36 &2.90 &59.58\% &24.22\% &74.48\% &22.19\% &77.19\% &93.41\% \\
&RS &8.2 &3.1 &60.6\% &88.1\% &71.8\% &88.2\% &71.2\% &92.3\% \\
% GS($\sigma=0.5$)  &15.53 &4.23 &95.41\% &8.07\% &38.02\% &7.81\% &42.19\% &92.53\% \\
&MART &9.3 &\textbf{3.3} &\textbf{67.8\%} &88.7\% &\textbf{76.6\%} &88.1\% &76\% &93.0\% \\
&TRADES &\textbf{9.4} &3.2 &62.3\% &89.4\% &76.3\% &88.7\% &75.4\% &93.0\% \\
&\textbf{MMAT} &8.8 &\textbf{3.3} &61.3\% &\textbf{89.6\%}  &75.9\%  & \textbf{89.1\%} &\textbf{76.3\%} & \textbf{93.2\%} \\
\hline
\end{tabular}}
\end{table}

% Both BASAR-MMAT and SMART-MMAT can improve the robustness (Table~\ref{tab:compar}) and SMART-MMAT has slightly better overall performance. This is understandable since SMART-MMAT is white-box and computes more aggressive adversaries. However, BASAR-MMAT can achieve better accuracy than standard training and SMART-MMAT, which show BASAR-MMAT can potentially eliminate the accuracy-robustness trade-off. Overall, MMAT is effective under both white-box and black-box samplers. 
\begin{table}[!tb]
\caption{Adaptive attack to evaluate the obfuscated gradient effect. ($\pm$xx\%) means the robustness difference with SMART-200.}
\label{tab:EOT}
\centering
\resizebox{0.6\linewidth}{!}{
\begin{tabular}{c|c|c|c}
\hline
SMART-MMAT & ST-GCN   &MS-G3D  &SGN  \\ \hline
HDM05 & 46.41\% (-0.4\%) &83.92\% (-0.9\%)  & 46.41\% (-2.0\%) \\ \hline
\end{tabular}}

%\vspace{-0.3cm}
\end{table}
\subsection{Gradient Obfuscation Evaluation}
Gradient obfuscation can potentially lead to failure in defense methods~\cite{athalye2018obfuscated}, because obfuscated gradients can be circumvented and are not truly robust. Considering that adaptive attack has become the de facto standard for evaluating gradient obfuscation~\cite{athalye2018obfuscated, tramer2020adaptive}, we following the adaptive attack criterion~\cite{tramer2020adaptive} to deploy an adaptive attack called EOT-SMART for skeleton-based HAR: in each step, we estimate the expected gradient by averaging the gradients of multiple randomly interpolated samples. Table \ref{tab:EOT} shows that the robustness under the adaptive attack only slightly worse than original SMART-200, demonstrating that MMAT does not rely on obfuscated gradients.

\section{Disscussion and Conclusion}
We proposed the very first black-box adversarial attack method which gives strong performance across datasets, models and attack modes. More broadly, we show, for the first time, the wide existence of on-manifold adversarial samples in skeletal motions. We also proposed a new adversarial training method to achieve simultaneous improvement on accuracy and robustness in general. One limitation is that BASAR relies on an explicit manifold parameterization which is not always feasible in other data types, e.g. videos~\cite{yang2021deconfounded}. This can be mitigated by learning from a large video dataset and use the learned model to do the manifold projection. Finally, BASAR adversarial samples can be theoretically realized by humans because they are on the natural manifold. However, how to attack a system in the real world using BASAR still depends on the specific setting of the system. In this research, we make the first step to identify the potential danger. In future, we will extend BASAR in other modalities via implicit manifold parameterization.

% One limitation is that BASAR relies on an explicit manifold parameterization which is not always feasible in other data types, e.g. videos~\cite{yang2021deconfounded}. This can be mitigated by learning from a large video dataset and use the learned model to do the manifold projection. Finally, BASAR adversarial samples can be theoretically realized by humans because they are on the natural manifold. 
%liu2023cross
\section*{Acknowledgment}
This project has received funding from the EU H2020 (No 899739), EPSRC (EP/R031193/1), NSF China (No. 62302139, No. 61772462, No. U1736217), RCUK grant CAMERA (EP/M023281/1, EP/T022523/1), FRFCU-HFUT (JZ2023HGTA0202, JZ2023HGQA0101).

%% The Appendices part is started with the command \appendix;
%% appendix sections are then done as normal sections
\appendix

\section{Additional Attack Experiments}
\label{Sec:Def}
\subsection{Detailed Settings}
We select five models: ST-GCN~\cite{Yan:2018:STG}, MS-G3D~\cite{Liu_2020_CVPR}, SGN~\cite{Zhang_2020_CVPR_1}, CTR-GCN~\cite{ctrgcn} and FR-HEAD~\cite{frhead}, and five benchmark datasets: HDM05~\cite{cg-2007-2}, NTU 60~\cite{shahroudy2016ntu}, Kinetics-400~\cite{kay2017kinetics}, UAV-Human~\cite{Li2021UAVHuman} and Skeletics-152~\cite{skeletics152} for experiments. HDM05~\cite{cg-2007-2} has 130 action classes, 2337 sequences from 5 subjects. Its high quality makes it suitable for our perceptual study, where any visual difference between the adversarial and the original motion is easily noticeable. We process HDM05 following~\cite{wang_understanding_2021} and train the target models achieving 87.2\%, 94.4\%, 94.1\%, 95.1\% and 92.3\% on ST-GCN, MS-G3D, SGN, CTR-GCN and FR-HEAD respectively. NTU60~\cite{shahroudy2016ntu} includes 56578 skeleton sequences with 60 action classes from 40 subjects. Due to the large intra-class and viewpoint variations, it is ideal for verifying the effectiveness and generalizability of our approach. Kinetics-400~\cite{kay2017kinetics} is a large and highly noisy dataset taken from different YouTube Videos. Similarly, Skeletics-152 is a 3D pose-annotated subset of videos sourced from much larger Kinetics-700 datasets~\cite{kinetics700}. UAV-Human~\cite{Li2021UAVHuman} is a challenging benchmark since it was collected by an unmanned aerial vehicle. We randomly split HDM05 datasets into training and test sets in a 7:3 ratio and attack the whole test set for evaluation. While for other large-scale datasets, sampling the whole test set is prohibitively slow, so we randomly sample motions to attack. We start with randomly sampling a small number of motions to attack and gradually increase the sample number until all evaluation metrics (explained below) stabilize, so that we know the attacked motions are sufficiently representative in the dataset. At the end, we end up with 1200 motions on NTU 60. Due to the large size of motion data in Kinetics-400, UAV-Human, and Skeletics-152, we randomly sample 500 motions from each dataset following the same sampling protocol. The experiments are conducted on a Xeon Silver 4216 CPU and an NVIDIA GTX 2080Ti GPU. The query time and number of queries are shown in Table~\ref{tab:query}.

\begin{table}[!htb]
\caption{The averaged number of queries and time (min) for generating an adversarial sample on different models and datasets.}
\label{tab:query}
\centering
\resizebox{0.6\linewidth}{!}{
\setlength{\tabcolsep}{1.2mm}{
\begin{tabular}{c|c|cc|cc|cc}
\hline
\multirow{2}{*}{Models} & \multirow{2}{*}{} & \multicolumn{2}{c|}{HDM05} & \multicolumn{2}{c|}{NTU} & \multicolumn{2}{c}{Kinetics-400} \\
                        &                   & Queries       & Time       & Queries      & Time      & Queries         & Time       \\ \hline
\multirow{2}{*}{ST-GCN}  & UA                & 3636          & 4          & 7337         & 12        & 7167            & 28          \\
                        & TA                & 8862          & 15         & 15724        & 16        & 15234           & 41          \\ \hline
\multirow{2}{*}{MS-G3D}  & UA                & 3722          & 6          & 14640        & 18        & 7190            & 29          \\
                        & TA                & 9111          & 16         & 23227        & 30        & 15416           & 56          \\ \hline
\multirow{2}{*}{SGN}    & UA                & 974          & 4         & 623         & 5         & 228             & 10          \\
                        & TA                & 277           & 3          & 260          & 4         & 180             & 8           \\ \hline
\end{tabular}}}
\end{table}

\subsection{Classifier Robustness}
\label{sec:robustness}
From the varying performances across target models and attack modes, we can see that classifiers are not equally gullible. SGN, in general, is the hardest to fool, requiring larger perturbation compared with ST-GCN and MS-G3D. This includes both joint-angle and joint-position attack. We speculate that it has to do with the features that SGN uses. Unlike ST-GCN and MS-G3D which use raw joints and bones and rely on networks to learn good features, SGN also employs \textit{semantic} features, where different joint types are encoded in learning their patterns and correlations. This requires large perturbations to bring the motion out of its pattern. Therefore, semantic information improves its robustness against attacks, as larger perturbations are more likely to be perceptible. Although SGN sometimes has a higher OM percentage, we find that some OM motions have noticeable differences from the original motions. In other words, unlike ST-GCN and MS-G3D where a higher OM percentage indicates more visually indistinguishable adversarial samples, some OM samples of SGN look natural, can fool the classifier and probably can fool humans when being observed independently, but are unlikely to survive strict side-by-side comparisons with the original motions. Next, MS-G3D is slightly harder to fool than ST-GCN. Although both use joint positions, MS-G3D explicitly uses the bone information, which essentially recognizes the relative movements of joints. The relative movement pattern of joints helps resist attack. 

% Finally, the robustness of a specific model varies across datasets. Meanwhile, the pattern of robustness variations across datasets also differs from model to model. It is hard to theoretically identify the cause and we will leave it to future research. 

To further verify our analysis, we employ an ablation study to investigate the relation between the number of queries and metrics, shown in Fig~\ref{fig:HDM05} to Fig~\ref{fig:KINETICS}. Being consistent with our analysis, compared with ST-GCN and MS-G3D, SGN usually converges faster but it is difficult for BASAR to further improve the adversarial sample as it does on ST-GCN and MS-G3D. We speculate that this is because SGN uses
semantic information, which prevents small perturbations from altering the class labels. More details can be found in the supplementary materials.

\section{Additional Defense Experiments}
\subsection{Ablation Studies}
\label{sec:as}
Since there are three weights in our adversarial training loss, we conduct an ablation study in different settings to evaluate their impacts, including (1) standard training ($\mu_{c} = 1$), (2) adversarial training ($\mu_{c} = 0$) (3) mixture training with on-manifold adversarial samples and clean samples ($\mu_{off} = 0$), (4) mixture training with off-manifold adversarial samples and clean samples ($\mu_{on} = 0$) (5) mixture training with on-manifold, off-manifold adversarial samples and clean samples. Table \ref{tab:defense} lists the defense results of BASAR-MMAT under different loss setting across different models. We observe BASAR-MMAT can simultaneously improve the accuracy and robustness of all three models. The accuracy even outperforms the prior arts in standard training ($+4\%$, $+1.5\%$, $+0.8\%$ on ST-GCN, MS-G3D, SGN respectively). This is also different from existing AT methods~\cite{bountakas2023defense}, which have to determine the trade-off between accuracy and robustness, BASAR-MMAT can achieve both goals. %MMAT can seem to achieve both targets with best performance without paying any price, or only at a tiny price. 
We speculate that this is because on-manifold adversarial samples can be essentially treated as unobserved ground truth data, which is strongly related to the generalization error~\cite{stutz_Disentangling_2019}. Therefore, incorporating such samples into training naturally boosts the accuracy. Meanwhile, they help move boundaries to include the surrounding areas of the data, and hence improve the robustness. 

The optimal weighting however varies across models. For ST-GCN and MS-G3D, the effects of on/off-manifold adversarial samples are consistent with our expectation, i.e. on-manifold adversarial examples mainly contributing to improving accuracy while off-manifold adversarial examples mainly contributing on improving robustness. Somewhat surprisingly, the results on SGN show that on-manifold samples alone can achieve both goals, and adding off-manifold samples could worsen the performance (the last and penultimate row in Table \ref{tab:defense}). We speculate that this is largely due to the structure and geometry of the decision boundaries of different models. Nevertheless, BASAR-MMAT can always achieve better accuracy and robustness. The ablation results of SMART-MMAT are shown in Table \ref{tab:SMART-MMAT}. Similarly, SMART-MMAT has consistent robustness performance across various tested attacks, and does not compromise standard accuracy.

\begin{table}[!t]
\caption{Robustness of BASAR-MMAT against BASAR-NoMP. Top: ST-GCN, Middle: MS-G3D, Bottom: SGN. Acc is the natural classification accuracy, i.e. average accuracy on clean data.} 
\label{tab:defense}
\centering
\resizebox{0.8\linewidth}{!}{
\begin{tabular}{c|ccc|c}
\hline
$\mu$ & $l\uparrow$ & $\Delta a\uparrow$ & $\Delta B/B\uparrow$ & Acc$\uparrow$ \\
\hline
$\mu_{c}=1$,\ $\mu_{on}=0$, \ $\mu_{off}=0$  & 0.10  & 0.04  & 0.66\% & 87.2\% \\

$\mu_{c}=0$,\ $\mu_{on}=0.4$, \ $\mu_{off}=0.6$  & 1.23  & 0.40  & 6.37\% &89.8\% \\

$\mu_{c}=0.4$,\ $\mu_{on}=0.6$, \ $\mu_{off}=0$  & 0.76  & 0.17  & 2.85\% & 90.0\% \\

$\mu_{c}=0.4$,\ $\mu_{on}=0$, \ $\mu_{off}=0.6$  & 1.28  & 0.30  & 5.03\% & 85.5\% \\

$\mu_{c}=0.4$,\ $\mu_{on}=0.2$, \ $\mu_{off}=0.4$  & \textbf{2.07}  & \textbf{0.67}  & \textbf{10.91\%} & \textbf{91.2\%} \\
\hline\hline
$\mu_{c}=1$,\ $\mu_{on}=0$, \ $\mu_{off}=0$  & 0.70  & 0.09  & 1.81\% & 94.4\% \\

$\mu_{c}=0$,\ $\mu_{on}=0.4$, \ $\mu_{off}=0.6$  & 4.17  & 0.79  & 14.22\% & 93.0\% \\

$\mu_{c}=0.6$,\ $\mu_{on}=0.4$, \ $\mu_{off}=0$  & 2.33  & 0.44  & 7.90\% & 93.6\% \\

$\mu_{c}=0.4$,\ $\mu_{on}=0$, \ $\mu_{off}=0.6$  & \textbf{4.53}  & \textbf{0.86}  & \textbf{15.86\%} & 94.4\% \\

$\mu_{c}=0.6$,\ $\mu_{on}=0.2$, \ $\mu_{off}=0.2$  & 4.09  & 0.79  & 14.15\% & \textbf{95.9\%} \\

\hline\hline
$\mu_{c}=1$,\ $\mu_{on}=0$, \ $\mu_{off}=0$  & 7.93  & 2.00  & 39.64\% & 94.1\% \\

% $\mu_{c}=0$,\ $\mu_{on}=0.4$, \ $\mu_{off}=0.6$  & 12.54  & 2.14  & 41.52\% & 91.1\% \\

$\mu_{c}=0.6$,\ $\mu_{on}=0$, \ $\mu_{off}=0.4$  & 10.97  & 1.60  & 29.39\% & 94.1\% \\

$\mu_{c}=0.4$,\ $\mu_{on}=0.6$, \ $\mu_{off}=0$  & 13.73  & 3.63  & 77.14\% & \textbf{94.9\%} \\

$\mu_{c}=0.6$,\ $\mu_{on}=0.4$, \ $\mu_{off}=0$  & \textbf{14.85}  & \textbf{3.76}  & \textbf{80.26\%} & 94.7\% \\
$\mu_{c}=0.4$,\ $\mu_{on}=0.4$, \ $\mu_{off}=0.2$  & 11.84  & 2.27  & 43.86\% & 93.9\% \\
\hline
\end{tabular}}
\end{table}

\begin{table}
\caption{Robustness of SMART-MMAT against white-box attacks with different iterations. Top: STGCN, Middle: MSG3D, Bottom: SGN. Acc is the classification accuracy, i.e. average accuracy on clean data.} 
\label{tab:SMART-MMAT}
\centering
\resizebox{1.0\linewidth}{!}{
\setlength{\tabcolsep}{1mm}{
\begin{tabular}{c|cc|cc|c}
\hline
\multirow{2}{*}{$\mu$}  &\multicolumn{2}{c|}{SMART}  &\multicolumn{2}{c|}{CIASA} &\multirow{2}{*}{Acc}\\
 & @50 & @200  & @50 & @200 &\\
\hline
 $\mu_{c}=1$,\ $\mu_{on}=0$, \ $\mu_{off}=0$  & 3.76\%  & 0.58\%  & 1.69\% &0.58\% & 87.2\% \\

$\mu_{c}=0$,\ $\mu_{on}=0.4$, \ $\mu_{off}=0.6$  & \textbf{86.05\%}  & \textbf{65.23\%}  & 84.85\%  &\textbf{61.63\%} &89.6\%\\

$\mu_{c}=0.4$,\ $\mu_{on}=0.6$, \ $\mu_{off}=0$  & 85.33\%  & 64.31\%  & 84.52\% &61.06\% & 90.6\% \\

$\mu_{c}=0.4$,\ $\mu_{on}=0$, \ $\mu_{off}=0.6$  & 81.16\%  & 62.83\%  & 81.88\% &59.59\% & 85.3\% \\

$\mu_{c}=0.2$,\ $\mu_{on}=0.6$, \ $\mu_{off}=0.2$  &85.68\%  & 62.33\%  & \textbf{85.31\%} &60.91\% & \textbf{91.0\%} \\
\hline\hline
$\mu_{c}=1$,\ $\mu_{on}=0$, \ $\mu_{off}=0$  & 6.59\%  & 1.47\%  & 6.10\% &4.19\% & 94.4\% \\

$\mu_{c}=0$,\ $\mu_{on}=0.4$, \ $\mu_{off}=0.6$  &91.69\%  & \textbf{84.79\%}  & 91.71\% &\textbf{85.06\%}&93.8\% \\

$\mu_{c}=0.6$,\ $\mu_{on}=0.4$, \ $\mu_{off}=0$  & 91.64\%  & 79.75\%  & \textbf{92.42\%} & 80.72\% & \textbf{95.1\%} \\

$\mu_{c}=0.4$,\ $\mu_{on}=0$, \ $\mu_{off}=0.6$  & \textbf{91.84\%}  & 83.86\%  & 91.33\% &81.06\% & 94.2\% \\

$\mu_{c}=0.4$,\ $\mu_{on}=0.2$, \ $\mu_{off}=0.4$  & 91.58\%  & 80.82\%  & 91.58\% &78.05\% & 94.5\% \\

\hline\hline
$\mu_{c}=1$,\ $\mu_{on}=0$, \ $\mu_{off}=0$  & 63.29\%  & 22.56\%  & 69.15\%  &22.56\% & 94.1\% \\

$\mu_{c}=0$,\ $\mu_{on}=0.4$, \ $\mu_{off}=0.6$  & 84.67\%  & 32.31\%  & 81.61\% &28.24\% & 93.1\% \\

$\mu_{c}=0.4$,\ $\mu_{on}=0$, \ $\mu_{off}=0.6$  & 81.00\%  & 39.17\%  & 83.20\% &33.98\% & 93.4\% \\

$\mu_{c}=0.4$,\ $\mu_{on}=0.6$, \ $\mu_{off}=0$  & 83.20\%  & 29.36\%  & 80.27\% &27.42\% & 94.0\% \\

$\mu_{c}=0.4$,\ $\mu_{on}=0.4$, \ $\mu_{off}=0.2$  &\textbf{87.36\%}  &\textbf{48.44\%}  & \textbf{87.00\%} &\textbf{45.15\%} & \textbf{93.9\%} \\
\hline
\end{tabular}}}
\end{table}

% \begin{table}
% \caption{Defense performance of Gaussian Smoothing on HDM05 dataset.} 
% \label{tab:gaussian}
% \centering
% %\setlength{\tabcolsep}{0.7mm}{
% \begin{tabular}{|c|c|ccc|c|}
% \hline
% Models   & $\sigma$  & $l\uparrow$ & $\Delta a\uparrow$ & $\Delta B/B\uparrow$ & Acc$\uparrow$      \\  \hline
% \multirow{4}{*}{STGCN} & 0.01 &0.61 &0.20 &4.13\% &91.9\% \\  
%                       & 0.05 &1.98 &0.68 &10.89\% &90.6\% \\  
%                       & 0.1  &2.80 &1.16 &19.72\% &90.4\% \\
%                       & 0.5  &9.86 &4.43 &94.93\% &89.7\% \\  \hline\hline
% \multirow{4}{*}{MSG3D} & 0.01 &2.16 &0.67 &11.09\% &94.4\% \\  
%                       & 0.05 &4.91 &1.85 &32.70\% &94.5\% \\  
%                       & 0.1  &6.90 &2.59 &48.08\% &94.4\% \\
%                       & 0.5  &13.60 &5.95 &132.30\% &93.0\% \\  \hline\hline
% \multirow{4}{*}{SGN}   & 0.01 &6.13 &1.10 &19.10\% &94.7\% \\  
%                       & 0.05 &12.36 &2.90 &59.58\% &94.1\% \\  
%                       & 0.1  &12.87 &3.07 &64.18\% &94.1\% \\
%                       & 0.5  &15.53 &4.23 &95.41\% &93.9\% \\  \hline
% \end{tabular}
% \end{table}

\begin{figure*}[tb]
    \centering
    \includegraphics[width=\linewidth]{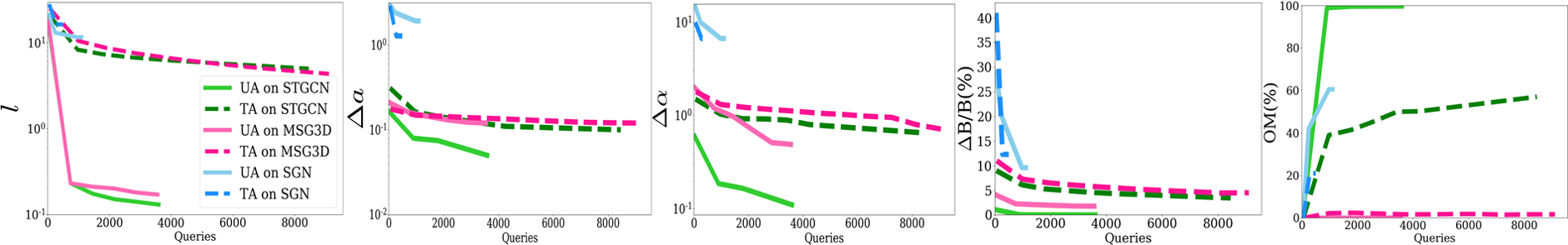}
    \caption{Metrics versus number of queries on HDM05 with STGCN, MSG3D and SGN. UA/TA refers to Untargeted Attack/Targeted Attack.}
    \label{fig:HDM05}
\end{figure*}
\begin{figure*}[tb]
    \centering
    \includegraphics[width=\linewidth]{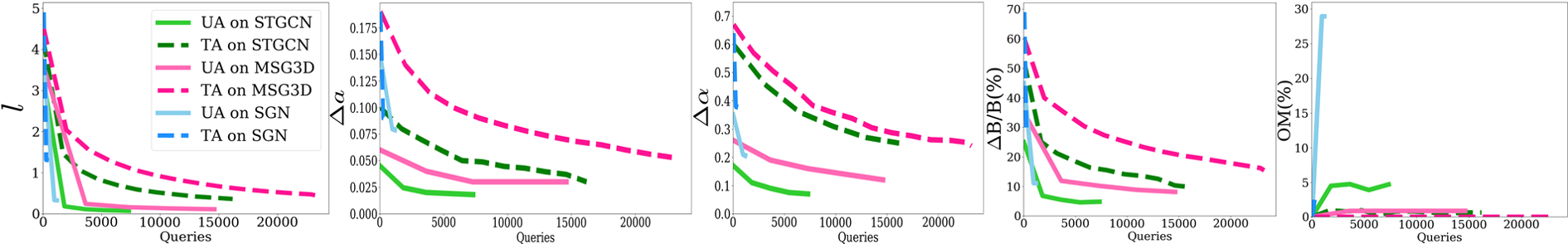}
    \caption{Metrics versus number of queries on NTU with STGCN, MSG3D and SGN.}
    \label{fig:NTU}
\end{figure*}
\begin{figure}[tb]
    \centering
    \includegraphics[width=\linewidth]{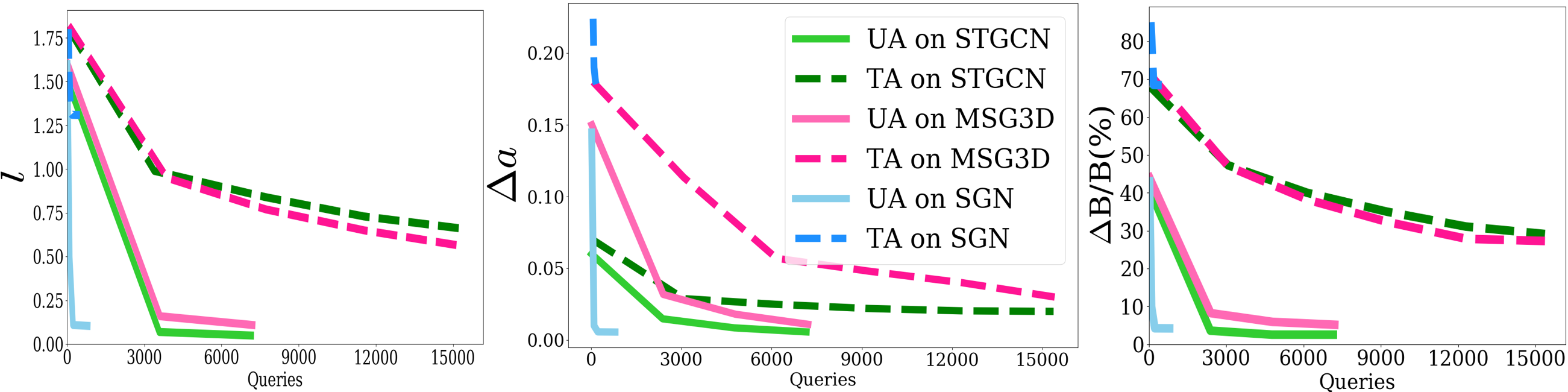}
    \caption{Metrics versus number of queries on Kinetics with STGCN, MSG3D and SGN.}
    \label{fig:KINETICS}
\end{figure}

\section{Perceptual Study}
One key difference between our work and existing work is that we employ both \textit{numerical accuracy} and rigorous \textit{perceptual studies} to evaluate the success of attacks. Imperceptibility is a requirement for any adversarial attack. All the success shown above would have been meaningless if the attack was noticeable to humans. To evaluate imperceptibility, rigorous perceptual studies are needed for complex data, as the numerical success can always be achieved by sacrificing the imperceptibility~\cite{wang_understanding_2021}. Therefore, we design a new perceptual study protocol that includes three perception metrics: Deceitfulness, Naturalness and Indistinguishability. Deceitfulness is to test whether BASAR visually changes the semantics of the motion. This is also to test whether people can distinguish actions by only observing skeletal motions. Naturalness aims to test whether on-manifold adversarial samples look more natural than off-manifold adversarial samples. Indistinguishability is the strictest test to see whether adversarial samples by BASAR can survive a side-by-side scrutiny. 

We specify the three perception metrics by conducting user studies. The details of the three user studies will be described below. Since we have 36 scenarios (models vs datasets vs attack types vs MP/No MP), it is unrealistic to exhaustively cover all conditions. We choose HDM05 and untargeted attack for our perceptual studies. We exclude NTU 60 and Kinetics-400 as they contain severe and noticeable noises. Our preliminary study shows that it is hard for people to tell if a motion is attacked therein. The quality of HDM05 is high where perturbations can be easily identified. In total, we recruited 50 subjects(ages between 20 and 54), of which 86\% of users are aged under 30 and 88\% are male. Approximately 25\% users have research expertise in HAR or adversarial attack; another 20\% have general deep learning or computer vision background; 45\% people study in engineering (e.g. mechanical, electrical). The other users have different arts background. By comparing their performance we found that the age, gender and work/research background do not have obvious correlations with the results. The results mainly depend on the quality of the adversarial samples.

% (ages between 20 and 54), of which 86\% of users are aged under 30 and 88\% are male. Approximately 25\% users have research expertise in HAR or adversarial attack; another 20\% have general deep learning or computer vision background; 45\% people study in engineering (e.g. mechanical, electrical). The other users have different arts background. By comparing their performance we found that the age, gender and work/research background do not have obvious correlations with the results. The results mainly depend on the quality of the adversarial samples.
%We briefly explain our study setting here and refer the readers to the supplementary materials for details.

\noindent\textbf{Deceitfulness}. In each user study, we randomly choose 45 motions (15 from ST-GCN, MS-G3D and SGN respectively) with the ground truth label and after-attack label for 45 trials. In each trial, the video is played for 6 seconds then the user is asked the question,`which label best describes the motion? and choose Left or Right', with no time limits. This is to test whether BASAR visually changes the meaning of the motion and whether the meaning of the original motion is clear to the subjects.
%We first test whether BASAR visually changes the meaning of the motion and whether the meaning of the original motion is clear to the subjects.

\noindent\textbf{Naturalness}. Since unnatural motions can be easily
identified as a result of the attack, we perform an ablation study to test whether on-manifold adversarial samples look more natural than off-manifold adversarial samples. We design two settings: MP and No MP. MP refers to BASAR, with Manifold Projection. No MP is where the proposed method without Manifold Projection. In each study, 60 (20 from ST-GCN, MS-G3D and SGN respectively) pairs of motions are randomly selected for 60 trials. Each trial includes one from MP and one from No MP. The two motions are played together for 6 seconds twice, then the user is asked, `which motion looks more natural? and choose Left, Right or Can't tell', with no time limits.

\noindent\textbf{Indistinguishability}. In this study, we conduct a very
strict test to see if the users can tell if a motion is perturbed in any way at all. In each user study, 40 pairs of motion are randomly selected, half from ST-GCN and half from MS-G3D. For each trial, two motions are displayed side by side. The left motion is always the original and the user is told so. The right one can be original (\textbf{sensitivity}) or attacked (\textbf{perceivability}). The two motions are played together for 6 seconds twice, then the user is asked, `Do they look same? and choose Yes or No', with no time limits. This user study serves two purposes. \textbf{Perceivability} is a direct test on \textbf{indistinguishability} while \textbf{sensitivity} aims to screen out users who tend to choose randomly. Most users are able to recognize if two motions are the same (close to 100\% accuracy), but there are a few whose choices are more random. We discard any user data which falls below 80\% accuracy on the sensitivity test.

\paragraph{Results}
The average success rate of \textbf{Deceitfulness} is 79.64\% across three models, with 88.13\% , 83.33\%, 67.47\% on ST-GCN, MS-G3D and SGN respectively. This is consistent with our prediction because SGN requires larger perturbations, thus is more likely to lead to the change of the motion semantics. Next, the results of \textbf{Naturalness} is that subjects regard 85\% on-manifold samples as more natural than their corresponding off-manifold samples. This is understandable as manifold projection not only makes sure the poses are on the manifold, but also enforces the similarity of the dynamics between the attacked and original motion. Finally, the results of \textbf{Indistinguishability} are 89.90\% on average. BASAR even outperforms the white-box attack (80.83\%) in \cite{wang_understanding_2021}. We further look into on-manifold vs off-manifold. Both samples are tested in Indistinguishability, but 94.63\% of the on-manifold samples fooled the users; while 84.69\% of the off-manifold samples fooled the users, showing that on-manifold samples are more deceitful. The off-manifold samples which successfully fool the users contain only small deviations.

\section{Comprehensive Visual Results and Confusion Matrices}

The visual results on various datasets and models are shown from Fig.~\ref{fig:visual_hdm05_stgcn} to Fig.~\ref{fig:visual_ntu_sgn}. As we can see, the adversarial samples on STGCN and MSG3D in general are very hard to be distinguished from the attacked motion. The results on SGN have the same semantic meanings and are almost equally hard to be distinguished from the original motion in untargeted attack. However, when it is targeted attack and the target label is very different from the original label, BASAR sometimes generate adversarial samples with visible differences. We show some failures here (Fig.~\ref{fig:visual_hdm05_sgn} Bottom and Fig.~\ref{fig:visual_ntu_sgn} Bottom.). These adversarial samples might survive a visual examination if shown alone but might not be able to survive a side-by-side comparison with the original motions in our rigorous perceptual studies. This is also consistent with our numerical evaluation.

In NTU, there are actions containing a single person or two persons. We, therefore, attack them separately. In targeted attack, if the attacked motion is a single-person action, the target class is also a single-person action where we randomly select a motion to initiate the attack. Similarly, if the attacked motion is a two-person action, we select a two-person motion. In untargeted attack, we do not need to initiate the attack separately and can rely on BASAR to find the adversarial sample that is closest to the original motion. The confusion matrices across various datasets and models are shown from Fig.~\ref{fig:confusion_hdm05_stgcn} to Fig.~\ref{fig:confusion_ntu_sgn}. 

In untargeted attack, we find that random attacks easily converge to a few action classes in a dataset. We call them \textit{high-connectivity classes}. For example, actions on STGCN tend to be attacked into `Jump Jack'(number 20) and `Kick left front'(21) on HDM05, and into `Use a fan'(48) on NTU, regardless how they are initialized; Similarly, actions on MSG3D tend to be attacked into `Cartwheel'(0) and `Kick right front'(23) on HDM05, and into `Use a fan'(48) on NTU; actions on SGN tend to be attacked into `Cartwheel'(0) and `Jump Jack'(20) on HDM05, and into `Hopping'(25) on NTU. The theoretical reason is hard to identify but we have the following speculations. Since untargeted attack starts from random motions, it is more likely to find the adversarial sample that is very close to the original motion on the classification boundary. Usually this adversarial sample is in a class that shares the boundary with the class of the original motion. It is possible that these high-connectivity classes share boundaries with many classes so that random attacks are more likely to land in these classes. In addition, the connectivity of classes heavily depends on the classifier itself and that is why different classifiers have different high-connectivity classes. In targeted attack, since our target labels are randomly selected, the confusion matrix is more uniformly distributed, covering all classes. 

More visual results can be found at: https://youtu.be/nF54LTa1KJo

\begin{figure}[!p]
    \centering
    \includegraphics[width=\linewidth]{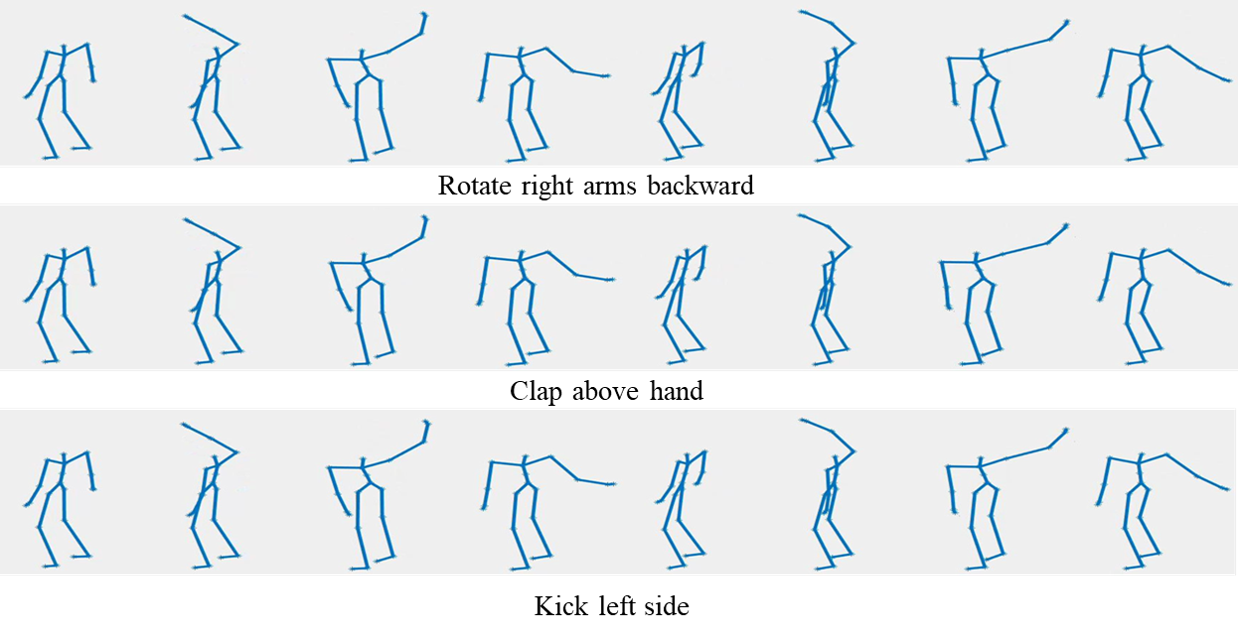}
    \caption{STGCN on HDM05. The ground truth label `Rotate right arms backward' is misclassified as `Clap above hand' on untargeted attack, and `Kick left side' on targeted attack.}
    \label{fig:visual_hdm05_stgcn}
\end{figure}
\begin{figure}[!p]
    \centering
    \includegraphics[width=\linewidth]{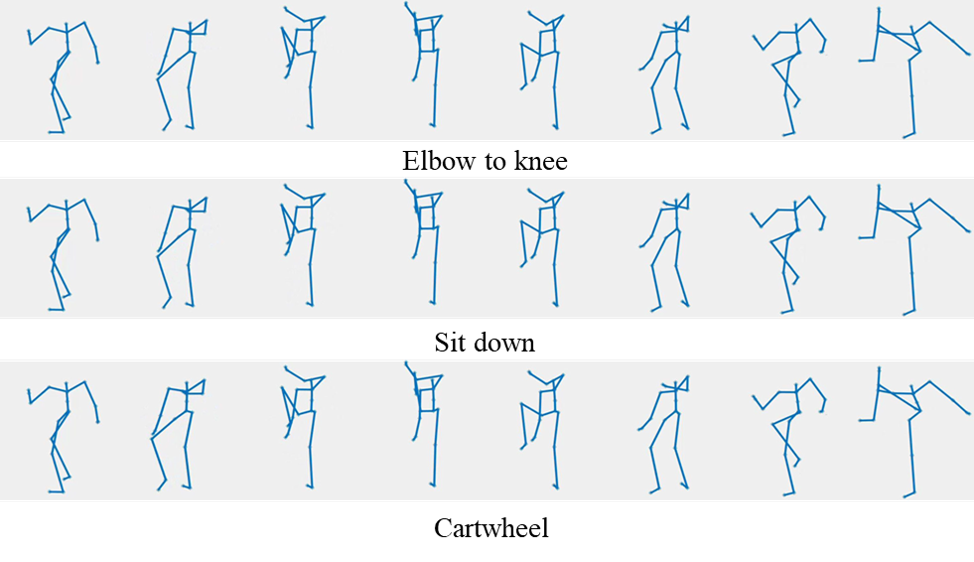}
    \caption{MSG3D on HDM05. The ground truth label `Elbow to knee' is misclassified as `Sit down' on untargeted attack, and `Cartwheel' on targeted attack.}
    \label{fig:visual_hdm05_msg3d}
\end{figure}
\begin{figure}[!p]
    \centering
    \includegraphics[width=\linewidth]{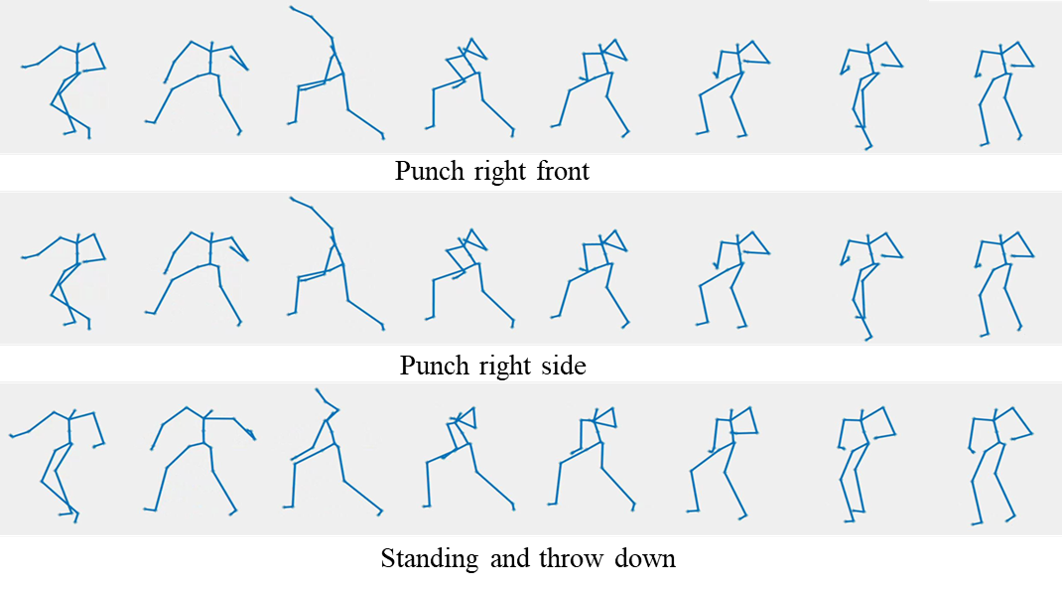}
    \caption{SGN on HDM05. The ground truth label `Punch right front' is misclassified as `Punch right side' on untargeted attack, and `Standing and throw down' on targeted attack.}
    \label{fig:visual_hdm05_sgn}
\end{figure}
\begin{figure}[!p]
    \centering
    \includegraphics[width=\linewidth]{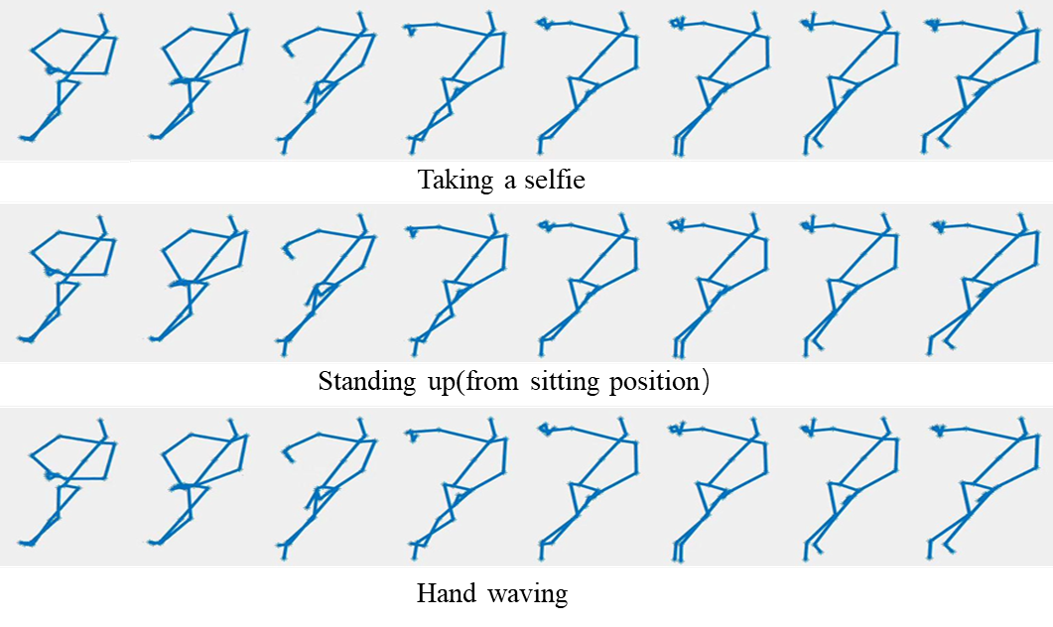}
    \caption{STGCN on NTU. The ground truth label `Taking a selfie' is misclassified as `Stand up' on untargeted attack, and `Hand waving' on targeted attack.}
    \label{fig:visual_ntu_stgcn}
\end{figure}
\begin{figure}[!p]
    \centering
    \includegraphics[width=\linewidth]{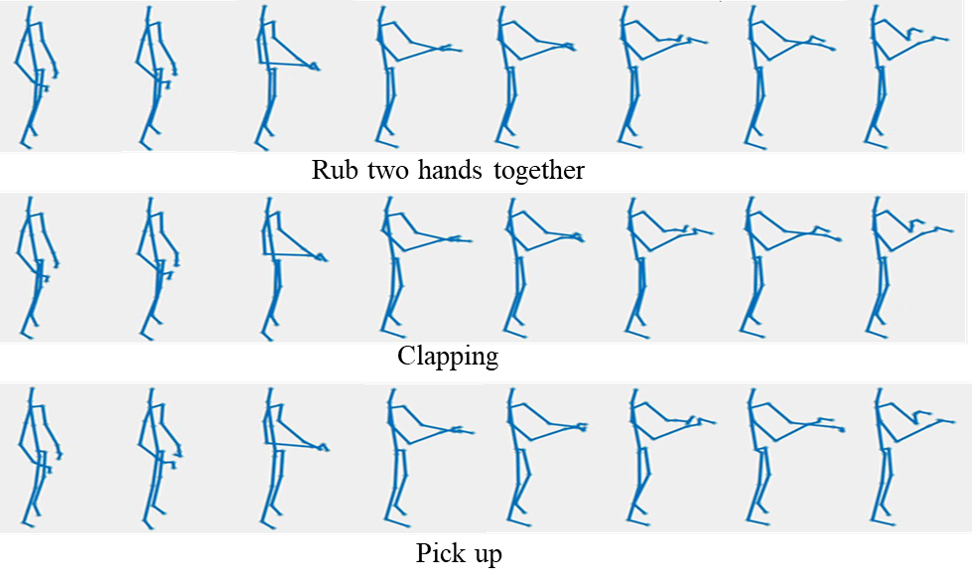}
    \caption{MSG3D on NTU. The ground truth label `Rub two hands together' is misclassified as `Clapping' on untargeted attack', and `Pick up' on targeted attack}
    \label{fig:visual_ntu_msg3d}
\end{figure}
\begin{figure}[!p]
    \centering
    \includegraphics[width=\linewidth]{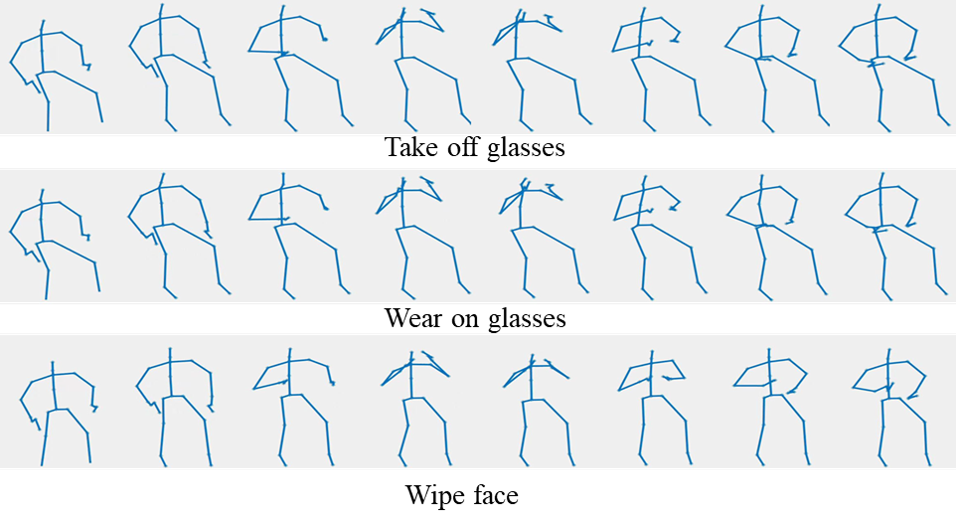}
        \caption{SGN on NTU. The ground truth label `Take off glasses' is misclassified as `Wear on glasses' on untargeted attack, and `Wipe face' on targeted attack. }
    \label{fig:visual_ntu_sgn}
\end{figure}

%%Consusion Matrix
\begin{figure}[!p]
    \centering
    \includegraphics[width=\linewidth]{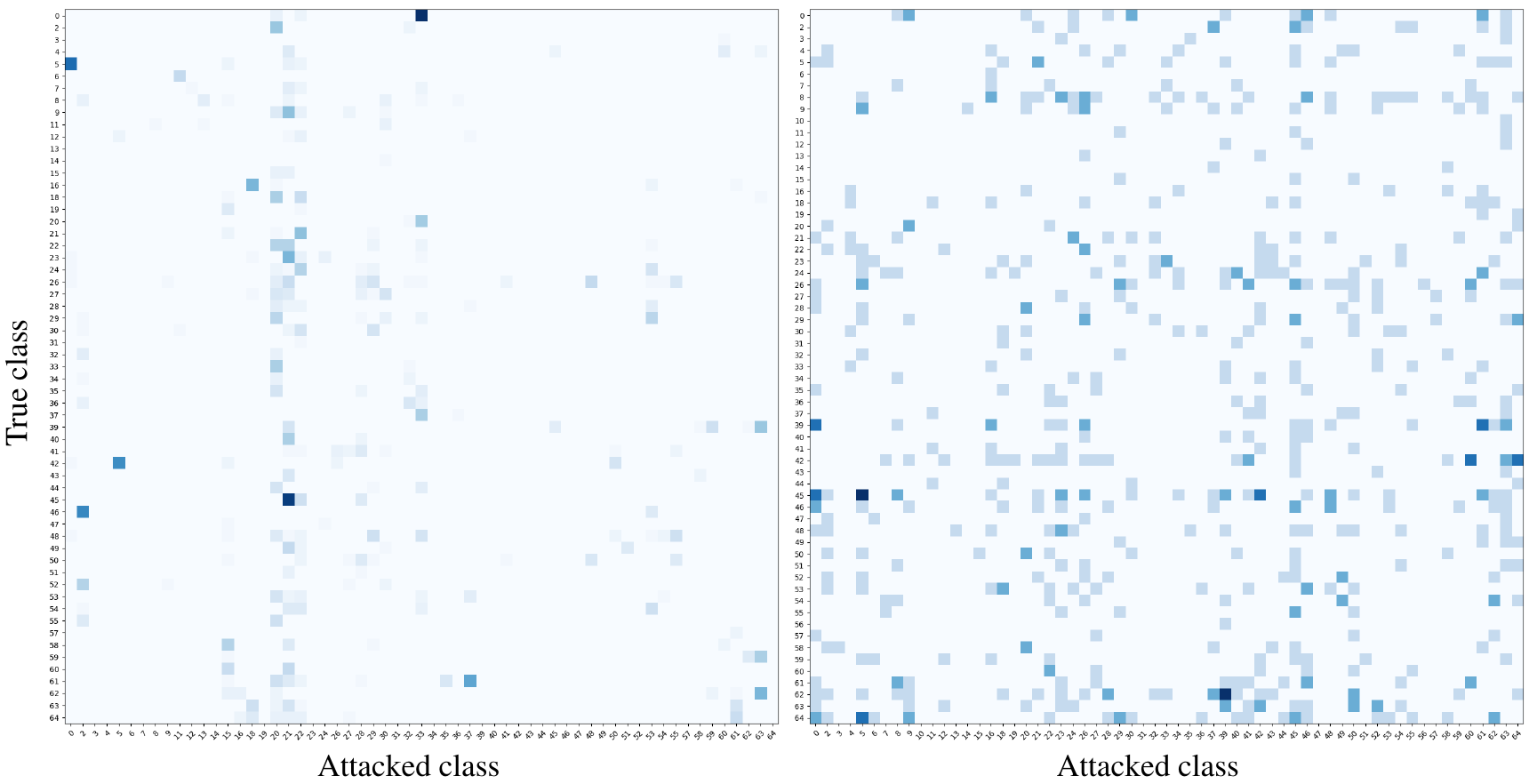}
    \caption{Confusion matrix of STGCN on HDM05. Left is untargeted attack and right is targeted attack. The darker the cell, the higher the value.}
    \label{fig:confusion_hdm05_stgcn}
\end{figure}
\begin{figure}[!p]
    \centering
    \includegraphics[width=\linewidth]{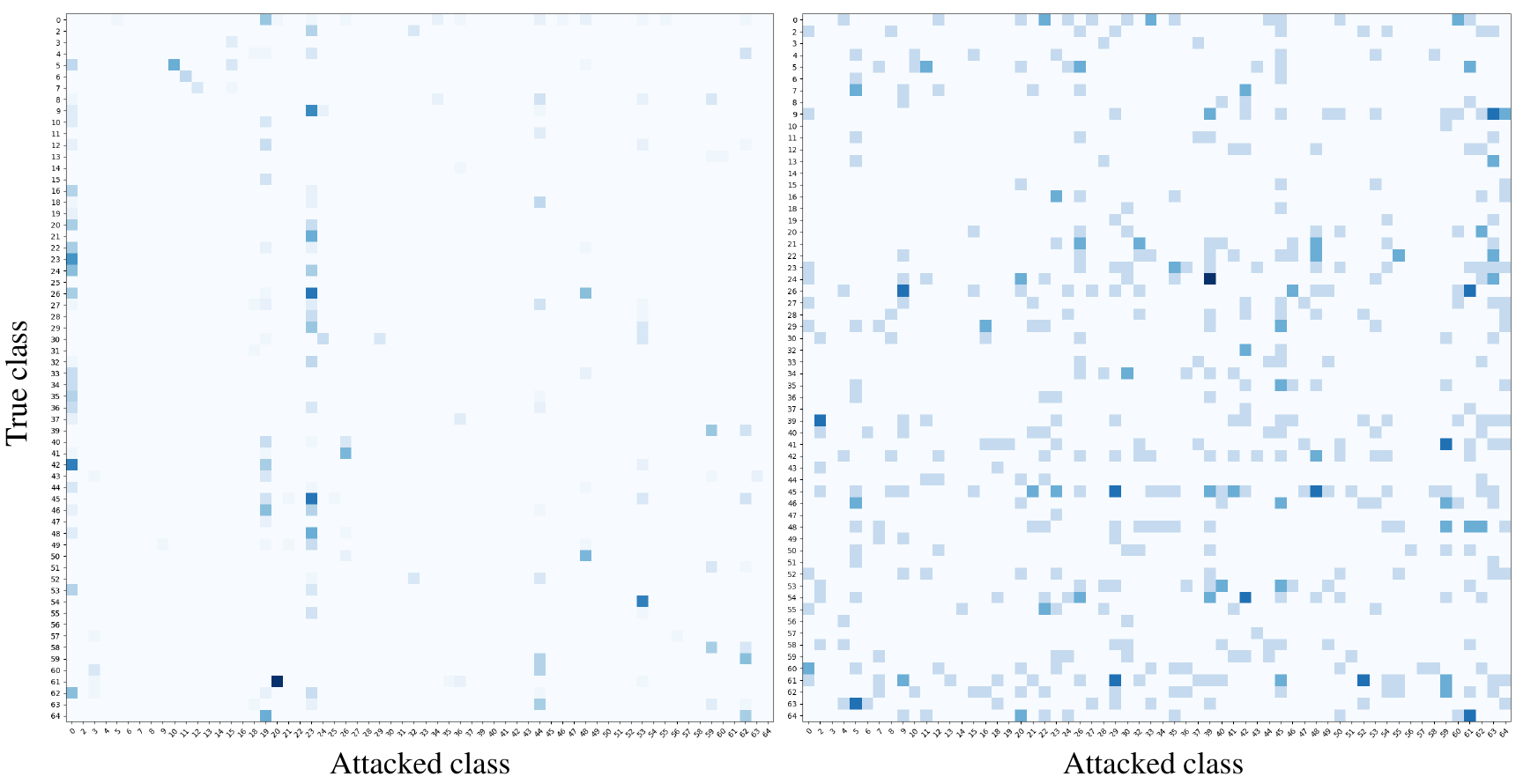}
    \caption{Confusion matrix of MSG3D on HDM05. The left one is untargeted attack and right is targeted attack. The darker the cell, the higher the value.}
    \label{fig:confusion_hdm05_msg3d}
\end{figure}
\begin{figure}[!p]
    \centering
    \includegraphics[width=\linewidth]{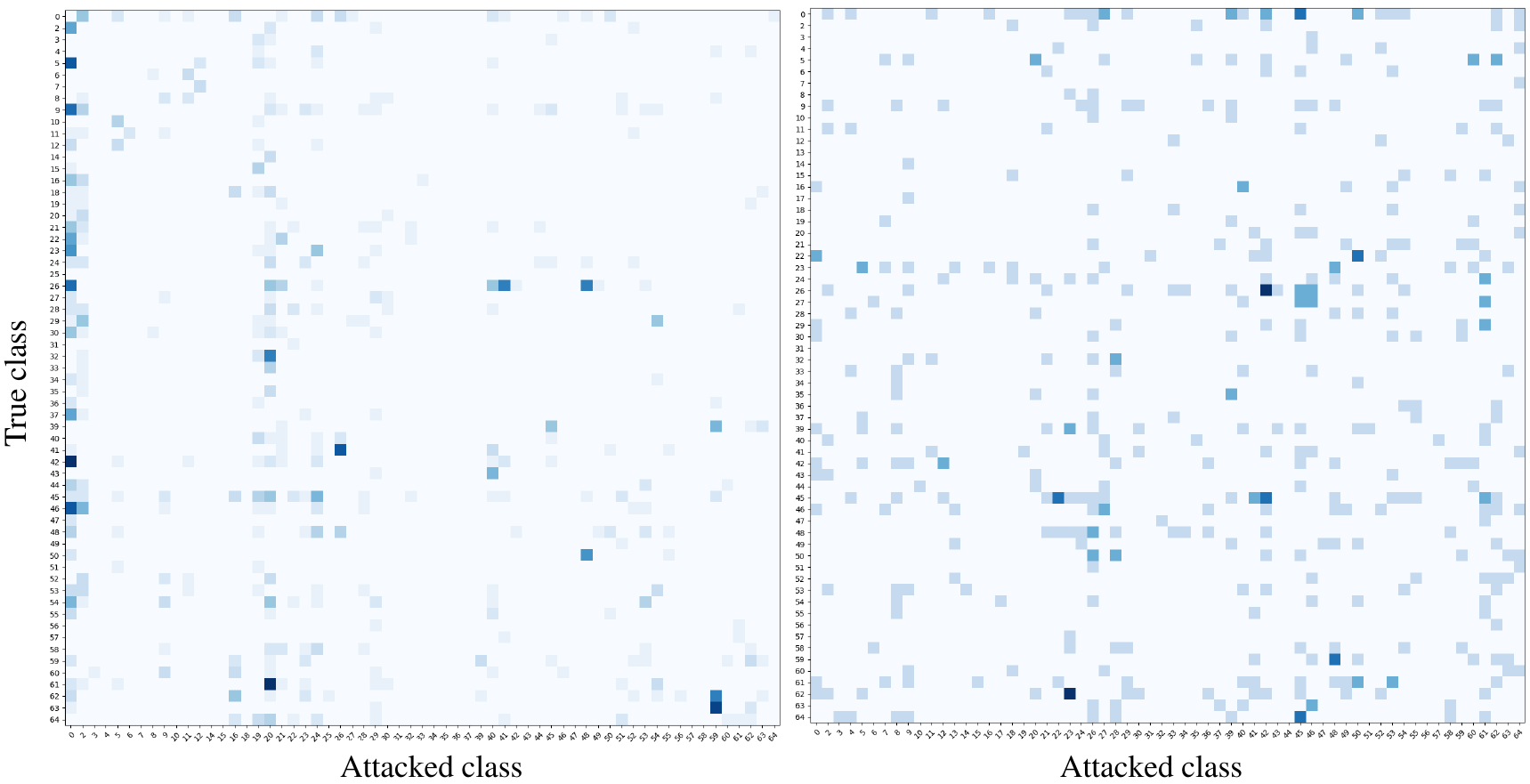}
    \caption{Confusion matrix of SGN on HDM05. The left one is untargeted attack and right is targeted attack. The darker the cell, the higher the value.}
    \label{fig:confusion_hdm05_sgn}
\end{figure}
\begin{figure}[!p]
    \centering
    \includegraphics[width=\linewidth]{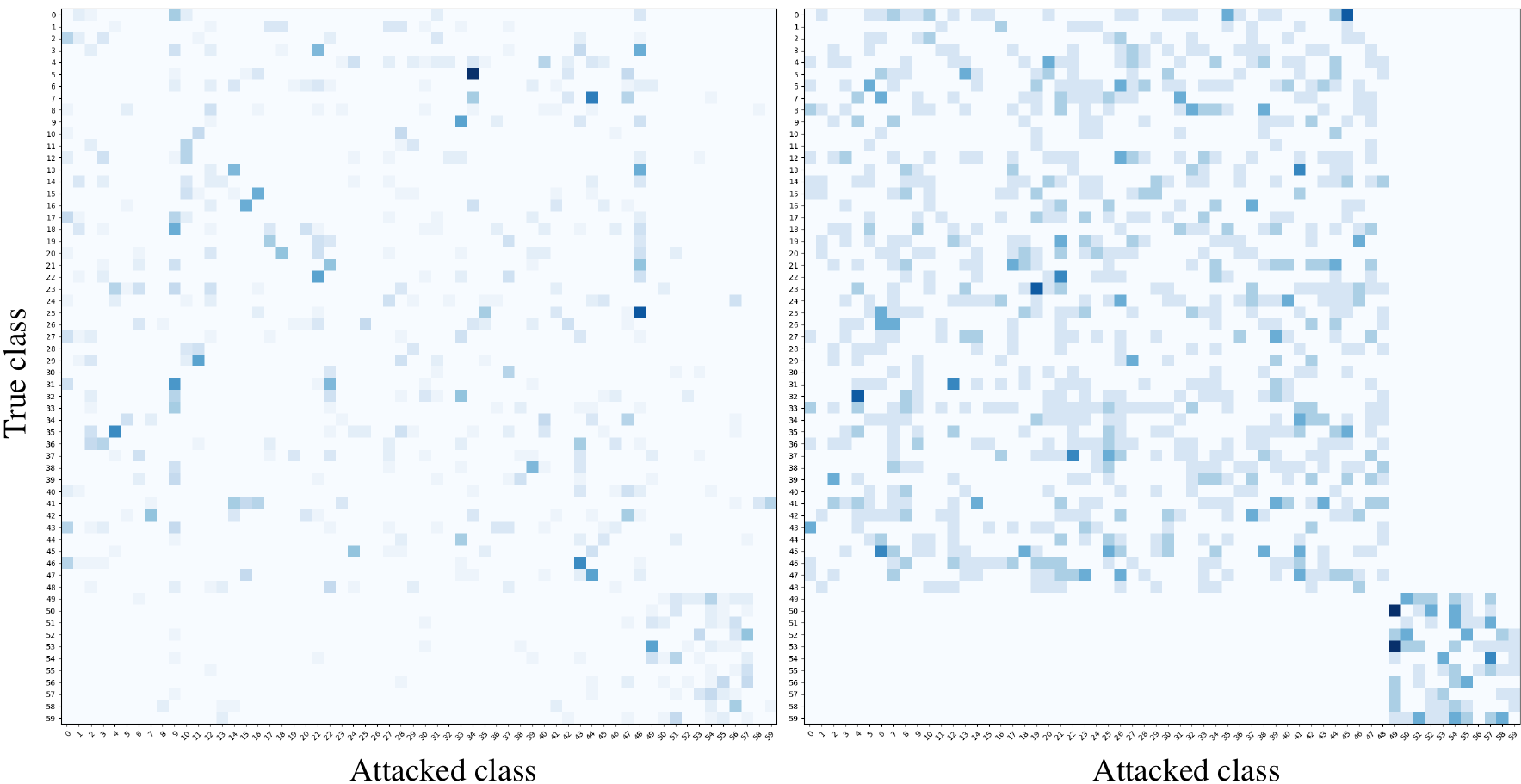}
    \caption{Confusion matrix of STGCN on NTU. The left one is untargeted attack and right is targeted attack. The darker the cell, the higher the value.}
    \label{fig:confusion_ntu_stgcn}
\end{figure}
\begin{figure}[tp]
    \centering
    \includegraphics[width=\linewidth]{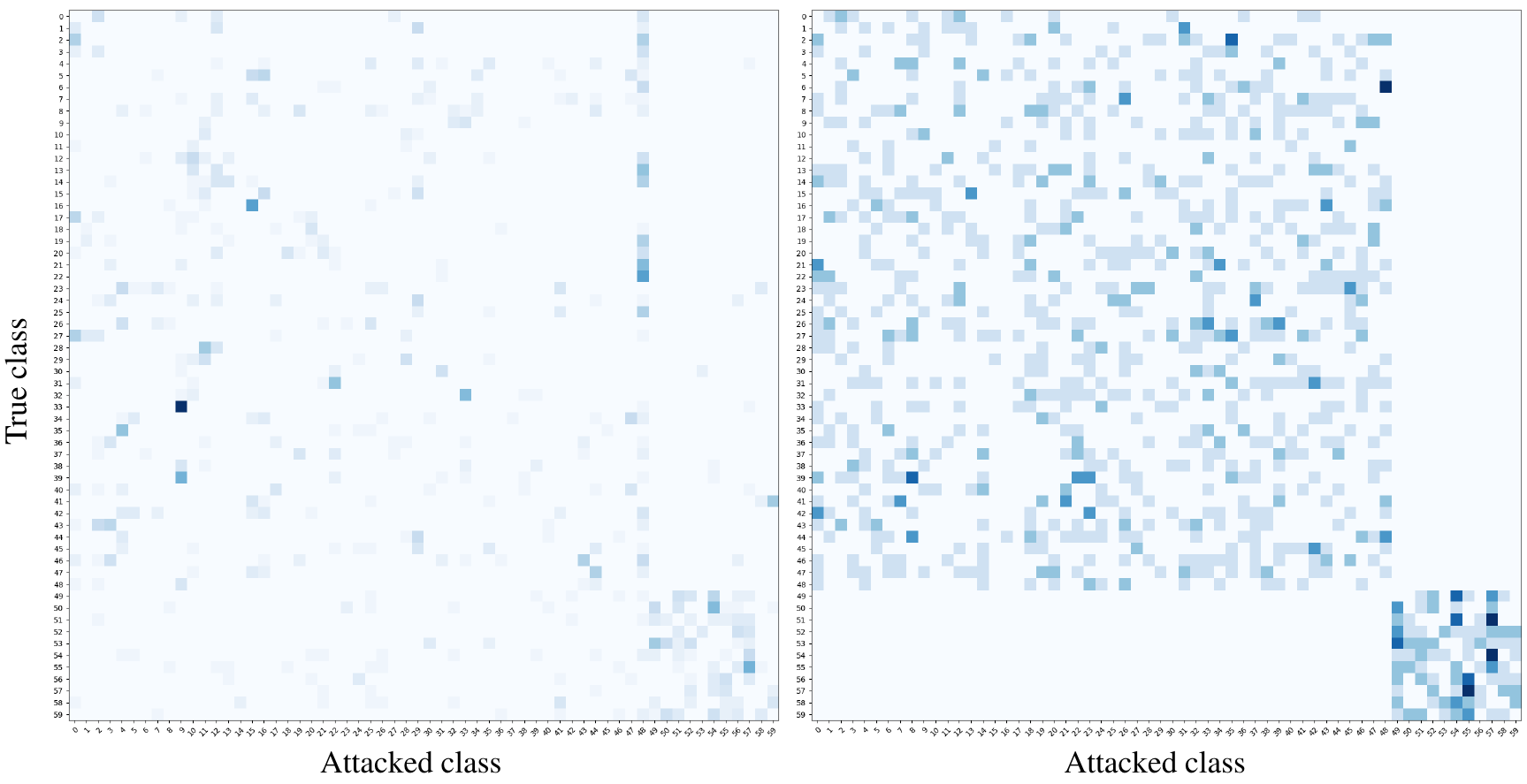}
    \caption{Confusion matrix of MSG3D on NTU. The left one is untargeted attack and right is targeted attack. The darker the cell, the higher the value.}
    \label{fig:confusion_ntu_msg3d}
\end{figure}
\begin{figure}[tp]
    \centering
    \includegraphics[width=\linewidth]{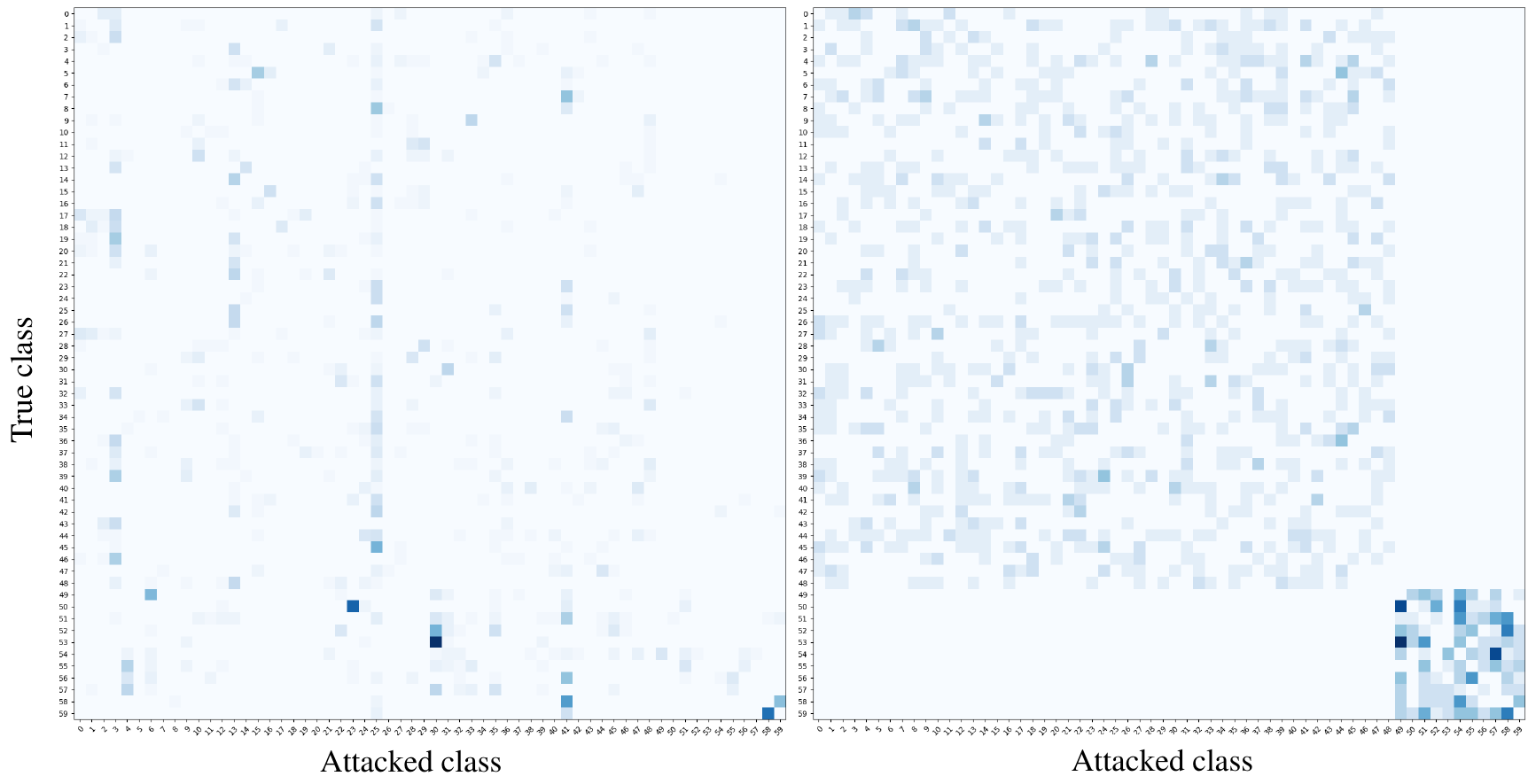}
    \caption{Confusion matrix of SGN on NTU. The left one is untargeted attack and right is targeted attack. The darker the cell, the higher the value.}
    \label{fig:confusion_ntu_sgn}
\end{figure}
% \section{Performances}
% The number of queries and time of BASAR are shown in Table~\ref{tab:experimental details}.
% \begin{table}[tb]
% \caption{The averaged number of queries and time (min) for generating an adversarial sample on different models and datasets.}
% \label{tab:experimental details}
% \centering
% \setlength{\tabcolsep}{1.2mm}{
% \begin{tabular}{c|c|cc|cc|cc}
% \hline
% \multirow{2}{*}{Models} & \multirow{2}{*}{} & \multicolumn{2}{c|}{HDM05} & \multicolumn{2}{c|}{NTU} & \multicolumn{2}{c}{Kinetics} \\
%                         &                   & Queries       & Time       & Queries      & Time      & Queries         & Time       \\ \hline
% \multirow{2}{*}{STGCN}  & UA                & 3636          & 4          & 7337         & 12        & 7167            & 28          \\
%                         & TA                & 8862          & 15          & 15724        & 16        & 15234           & 41          \\ \hline
% \multirow{2}{*}{MSG3D}  & UA                & 3722          & 6          & 14640        & 18        & 7190            & 29          \\
%                         & TA                & 9111          & 16         & 23227        & 30        & 15416           & 56          \\ \hline
% \multirow{2}{*}{SGN}    & UA                & 974           & 4          & 623          & 5         & 228             & 10           \\
%                         & TA                & 277           & 3          & 260          & 4         & 180             & 8          \\ \hline
% \end{tabular}}
% \end{table}
\section{Implementation details and mathematical deduction}
\label{Sec:Imp}
\textbf{Setting}. In Algorithm 1, the initial $\beta_1$ and $\beta_2$ are both set to 0.95. The initial $\lambda$ is set to 0.2 when attacking SGN model and 0.1 on both STGCN and MSG3D. We set the spinal joint weights to 0 in $\mathbf{W}$, and other joint weights to 1.  For untargeted attack, we set $\epsilon = 0.1$ on both HDM05 and NTU, 0.05 on Kinetics. For targeted attack, $\epsilon$ is set to 0.5 on HDM05 and both 0.2 on NTU and Kinetics. Considering the optimization speed, it is unrealistic to execute the manifold projection in every iteration. We therefore execute it every 100 iterations on HDM05 and every 250 iterations on NTU and Kinetics. The adversarial samples are computed using PyTorch on a PC with an NVIDIA GTX 2080Ti GPU and a Xeon Silver 4216 CPU. 

\noindent\textbf{Random Exploration}. For easy reference, the random exploration is reformulated in Eq.~\ref{eq:re}:
\begin{align}
\label{eq:re}
& \widetilde{\mathbf{x}} =\mathbf{x}' + \mathbf{W}\Delta, \nonumber \\
& \text{where }\Delta = \mathbf{R} - (\mathbf{R}^T\mathbf{d})\mathbf{d}, \  \mathbf{d} = \frac{\mathbf{x} - \mathbf{x}'}{\| \mathbf{x}- \mathbf{x}'\|}, \nonumber \\
& \mathbf{R} = \lambda \frac{\mathbf{r}}{\| \mathbf{r}\|} \| \mathbf{x} - \mathbf{x}'\|\text{, } r \in N(0, \mathbf{I}),
\end{align}
where $\widetilde{\mathbf{x}}$ is the new perturbed sample, $\mathbf{x}$ and $\mathbf{x}'$ are the attacked motion and current adversarial sample. The perturbation on $\mathbf{x}'$, $\Delta$, is weighted by $\mathbf{W}$ - a diagonal matrix with joint weights. $\Delta$ controls the direction and magnitude of the perturbation, and depends on two variables $\mathbf{R}$ and $\mathbf{d}$. $\mathbf{d}$ is the directional vector from $\mathbf{x}'$ to $\mathbf{x}$. $\mathbf{R}$ is a random directional vector sampled from a Normal distribution $N(0, \mathbf{I})$ where $\mathbf{I}$ is an identity matrix, $\mathbf{I} \in R^{z \times z}$, $z = mn$, $m$ is the number of Dofs in one frame and $n$ is total frame number. This directional vector is scaled by $\| \mathbf{x} - \mathbf{x}'\|$ and $\lambda$.

In the Random Exploration, we aim to find an adversarial sample that is closer to $\mathbf{x}$. However, as the shape of the local space is unknown and highly nonlinear, we do sampling to exploit it. Therefore, we execute multiple random explorations instead of only one to get $q$ intermediate results in a sub-routine call, and compute the attack success rate. If the rate is less than 40\%, $\lambda$ is reduced by 10\% as it means that we are very close to the classification boundary $\partial C$ and $\lambda$ is too big; if it is higher than 60\%, $\lambda$ is increased by 10\%; otherwise we do not update $\lambda$. 

For targeted attack, we randomly select one adversarial sample from the $q$ intermediate samples to do aimed probing. This is mainly to ensure that the direction of the aimed probing is random. Although multiple samples can be selected, it would incur more computational costs with little gain shown by our preliminary experiments. For untargeted attack, the $q$ results are normally in different classes which we call \textit{adversarial classes}. The attack difficulty varies depending on the choice of samples. Usually the closer the adversarial sample is to the original sample, the easier the attack. Therefore, different from~\cite{drendel_decisionaa_2018}, we randomly select one sample in each adversarial class to conduct aimed probing, then only keep the one that has the smallest distance to the original motion $\mathbf{x}$ after the aimed probing. In the end, when the adversarial sample is near to the original motion, we set a threshold value $\tau$ to ensure that $\lambda$ is not higher than $\tau$. This is to ensure that the attack can eventually converge. Empirically, in all experiments, We set $q=5$. $\tau$ is set to 1.5 on SGN and 0.4 on both STGCN and MSG3D.

%\noindent\textbf{Attack Strategy}. In NTU dataset, there are actions containing a single person or two persons. We, therefore, attack them separately. In targeted attack, if the attacked motion is a single-person motion, the target class is also a single-person motion where we randomly select a motion to initiate the attack. Similarly, if the attacked motion is a two-person motion, we select a two-person motion. In untargeted attack, we do not need to initiate the attack separately and can rely on BASAR to find the adversarial sample that is closest to the original motion. A confusion matrix is shown in Figure~\ref{fig:confusion_ntu_stgcn}. 

\noindent\textbf{Detailes of Manifold Projection}. The original problem is as follows:
\begin{align}
\label{eq:gmw_angles}
    & \min_{\mathbf{\theta'}}\ \ L(\widetilde{\mathbf{\theta}}, \mathbf{\theta'})  + w L(\mathbf{\ddot \theta}, \mathbf{\ddot \theta'})\ \ \text{s.t.}\ \ \theta_j^{\min} \le \mathbf{\theta'_j} \le \theta_j^{\max}, \nonumber \\
    & C_{\mathbf{x'}} = c\text{ (targeted)} \text{ or } C_{\mathbf{x'}} \ne C_{\mathbf{x}}\text{ (untargeted)}.
\end{align}
where $\mathbf{\theta}$ and $\widetilde{\mathbf{\theta}}$ are the joint angles of the original motion $\mathbf{x}$ and perturbed motion $\widetilde{\mathbf{x}}$, $\mathbf{\theta}'$ is the optimization variable, $\theta'_j$ is the $j$-th joint angle in every frame of $\mathbf{\theta}'$ and subject to joint limits bounded by $\theta_j^{\min}$ and $\theta_j^{\max}$. $\mathbf{\ddot \theta}$ and $\mathbf{\ddot \theta'}$ are the $2nd$-order derivatives of $\mathbf{\theta}$ and $\mathbf{\theta}'$. $w$ is a weight. $L$ is the Euclidean distance. 

We use a primal-dual interior-point method \cite{hedengren2014nonlinear} to solve Equation~\ref{eq:gmw_angles} which can be transformed to a barrier problem~\cite{conn2000primal}:
\begin{align}
\label{barrier}
    \min_{\mathbf{\theta'}} \ L(\widetilde{\mathbf{\theta}}, \mathbf{\theta'})  + w L(\mathbf{\ddot \theta}, \mathbf{\ddot \theta'}) +  \sum_{j}^{O} \nu_j \ln  {(\mathbf{\theta}'_j - \mathbf{\theta}_j^{min})} \nonumber \\
    +  \sum_{j}^{O} \nu_j \ln{(\mathbf{\theta}_j^{max} - \mathbf{\theta'}_j)}
\end{align}
where $\nu_j$ is a barrier parameter. $O$ is the total number of joints in a skeleton. For simplicity, we denote $f(\mathbf{\theta'}) = L(\widetilde{\mathbf{\theta}}, \mathbf{\theta'})  + w L(\mathbf{\ddot \theta}, \mathbf{\ddot \theta'})$. The Karush-Kuhn-Tucker conditions~\cite{kuhn2014nonlinear} for the barrier problem in Equation~\ref{barrier} can be written as:
\begin{align}
\label{eq:kkt}
    \nabla f(\mathbf{\theta'}) + \sum_{j}^{O} \frac{\nu_j}{\mathbf{\theta}'_j - \mathbf{\theta}_j^{min}} - \sum_{j}^{O} \frac{\nu_j}{\mathbf{\theta}_j^{max} - \mathbf{\theta}'_j}   = 0 \nonumber \\ 
    \nu_j >= 0, \text{ for } j = 1,..., O \nonumber, \ \ 
    \sum_{j}^{O} \nu_j \ln  {(\mathbf{\theta}'_j - \mathbf{\theta}_j^{min})} = 0 \nonumber \\
    \sum_{j}^{O} \nu_j \ln{(\mathbf{\theta}_j^{max} - \mathbf{\theta}'_j)} = 0
  % XZe - \mu e =0
\end{align}
We apply a damped Newton's method\cite{ypma1995historical} to compute an approximate solution to Equation \ref{eq:kkt}. More implementation details about the primal-dual interior-point method can be found in \cite{wachter2006implementation}. After solving for $\theta'$, the joint positions of the adversarial motion are computed using Forward Kinematics.

%% If you have bibdatabase file and want bibtex to generate the
%% bibitems, please use
%%
\bibliographystyle{elsarticle-num} 
\bibliography{Manuscript}

%% else use the following coding to input the bibitems directly in the
%% TeX file.

% \begin{thebibliography}{00}

% %% \bibitem{label}
% %% Text of bibliographic item

% \bibitem{}

% \end{thebibliography}
\end{document}